\documentclass{article} 
\usepackage{iclr2025_conference,times}
\definecolor{mydarkblue}{RGB}{0, 0, 139}

\usepackage{hyperref}
\usepackage{multirow}
\usepackage{url}
\usepackage{lineno}  

\usepackage{graphicx}
\usepackage{amsmath}
\usepackage{cleveref}
\usepackage{amssymb}
\usepackage{booktabs}
\usepackage{csquotes}
\usepackage{graphicx}  
\usepackage[numbers]{natbib}
\usepackage{subcaption}  
\usepackage{algorithm}
\usepackage{algpseudocode}

\usepackage{graphicx}    
\usepackage{subcaption} 
\hypersetup{colorlinks=true, citecolor=mydarkblue,linkcolor=mydarkblue}
\usepackage{longtable}

\begin{document}

\title{Scalable Robust Bayesian Co-Clustering with Compositional ELBOs
}

\author{Ashwin Vinod\textsuperscript{$^{\dagger}$},~
    Chandrajit Bajaj\textsuperscript{$^{\dagger}$}\\[0.5em]
    \textsuperscript{$^{\dagger}$}The University of Texas at Austin, Texas, USA
}

\footnotetext[1]{\textsuperscript{$^{\dagger}$}ashwinv@utexas.edu, bajaj@cs.utexas.edu}

\newcommand{\fix}{\marginpar{FIX}}
\newcommand{\new}{\marginpar{NEW}}

\iclrfinalcopy %






\maketitle

\begin{abstract}
Co‐clustering exploits the duality of instances and features to simultaneously uncover meaningful groups in both dimensions, often outperforming traditional clustering in high‐dimensional or sparse data settings. Although recent deep learning approaches successfully integrate feature learning and cluster assignment, they remain susceptible to noise and can suffer from posterior collapse within standard autoencoders. In this paper, we present the first fully variational Co‐clustering framework that directly learns row and column clusters in the latent space, leveraging a doubly reparameterized ELBO to improve gradient signal‐to‐noise separation. Our unsupervised model integrates a Variational Deep Embedding with a Gaussian Mixture Model (GMM) prior for both instances and features, providing a built‐in clustering mechanism that naturally aligns latent modes with row and column clusters. Furthermore, our regularized end-to-end noise learning Compositional ELBO architecture jointly reconstructs the data while regularizing against noise through the KL divergence, thus gracefully handling corrupted or missing inputs in a single training pipeline. To counteract posterior collapse, we introduce a scale modification that increases the encoder's latent means only in the reconstruction pathway, preserving richer latent representations without inflating the KL term. Finally, a mutual information-based cross-loss ensures coherent co-clustering of rows and columns. Empirical results on diverse real‐world datasets from multiple modalities, numerical, textual, and image-based, demonstrate that our method not only preserves the advantages of prior Co-clustering approaches but also exceeds them in accuracy and robustness, particularly in high‐dimensional or noisy settings.
\end{abstract}

\section{Introduction}
Co‐clustering aims to identify homogeneous clusters of both instances (rows) and features (columns) in a target matrix. This task has found widespread use in domains such as bioinformatics, text analysis, e-Commerce, and more. Over the years, numerous clustering methods have been proposed; however, most rely on linear transformations or shallow factorizations, which often fail to capture the non‐linear relationships found in real‐world data. In addition, these methods typically struggle with sparse, high-dimensional data and do not learn representations that can be readily used for downstream tasks. Traditional co‐clustering techniques also have difficulty handling missing entries, corrupted signals, and outliers.

To address these issues, we introduce a Scalable Bayesian Co-Clustering framework. Our approach uses a Variational Autoencoder (VAE) to capture non-linear relationships and discover more complex latent structures among both rows and columns. We further impose a Gaussian mixture model prior to regularizing the latent embeddings, helping to mitigate overfitting in the presence of noise or missing data. This deep learning–based framework is flexible enough to handle multiple data modalities and is also highly scalable for large datasets. In addition, its unified structure makes it easy to integrate additional objectives, enabling a versatile solution for modern clustering challenges. We also use the Scale trick from \cite{song2024scale} to alleviate posterior collapse and promote meaningful separation of latent variables, and stabilize joint learning of reconstruction and clustering objectives. 
\section{Related Work}
Various methods of co-clustering have been developed in the past. One such approach, proposed by \cite{dhillon2001co}, models the rows and columns of the data matrix as vertices of a bipartite graph and uses a normalized cut (via eigendecomposition or SVD) to find a partition that simultaneously clusters rows and columns. However, a major disadvantage of this framework is that it requires the same number of row and column clusters, and computing the eigen-decomposition for large datasets becomes computationally expensive. Another approach based on eigendecomposition is presented by \cite{kluger2003spectral}, which first normalizes the data matrix, then applies SVD to find checkerboard structures and finally identifies row/column clusters. However, this model is specifically designed for gene expression data and does not translate well to sparse data sets.

Other methods include the use of nonnegative matrix trifactorization (NMTF) \cite{ortho1}\cite{ortho2}\cite{ortho3}, which factorizes a nonnegative data matrix into three low-rank factors, imposing nonnegativity (and often orthogonality) constraints to induce simultaneous clustering of rows and columns. Unfortunately, these methods can be sensitive to initialization \cite{ortho4}\cite{ortho5}, can converge to local minima, and can be susceptible to noise. Extensions to NMTF involve adding graph (Laplacian) regularization (or manifold regularization) \cite{extend1}\cite{extend2} to preserve the geometric or manifold structure of rows/columns, but these approaches incur additional computational overhead from building and tuning the required graphs/manifolds. Information-theoretic methods \cite{infot} iteratively maximize mutual information $I(R,C)$ between a row partition $R$ and a column partition $C$. However, mutual information is unbounded in such approaches and typically favors more clusters, rendering parameter selection non-trivial.

Modularity-based co-clustering \cite{mod} adapts Newman's modularity \cite{newman} (originally from graph clustering) to a 'diagonal' co-clustering (using the same $k$ for rows and columns). However, this approach only provides diagonal solutions. Model-based co-clustering \cite{model1}\cite{model2} assumes that the data come from a mixture of distributions, one for each 'block' of rows-cluster-columns. These methods often use EM-like or variational procedures to learn the cluster memberships and model parameters, but they depend on correct distributional assumptions and can be slow for data of high dimensions or sparse \cite{models}. The ensemble co-clustering methods \cite{ensemble1}\cite{ensemble2} combine multiple co-clustering solutions (possibly from different algorithms or hyperparameter settings) into a consensus partition. They often rely on graph-based representations, with rows and columns as nodes and edges weighted by their frequency of co-occurrence in the same cluster. However, the decision of how to aggregate multiple solutions remains nontrivial. \cite{ghasedi2017deep} presents an end-to-end clustering framework based on a deep auto-encoder, which maps data to a discriminative embedding subspace and predicts cluster assignments. This approach has shown significantly strong performance on image datasets. One of the recent co-clustering methods that used deep learning includes Deep Co-Clustering \cite{deepcc} that employs auto-encoders to learn low-dimensional representations for instances and features, enabling co-clustering through mutual information maximization. Although effective in many cases, this approach is inherently sensitive to noisy data. Autoencoders minimize reconstruction loss, but lack robust mechanisms to disentangle noise from a meaningful structure in the latent space.  The reliance on standalone Gaussian Mixture Models (GMMs) for clustering can lead to suboptimal alignment between latent representations and cluster priors, particularly when noise disrupts learned embeddings. However, all of these methods differ from our variational clustering approach, which jointly learns structured latent representations for both instances and features within a single variational framework that is resilient to noisy and sparse inputs.

\section{Problem Formulation}

Let \(\{\mathbf{x}_i\}_{i=1}^{n}\) denote \(n\) instance (row) vectors and \(\{\mathbf{y}_j\}_{j=1}^{d}\) denote \(d\) feature (column) vectors. We wish to group instances into \(g\) clusters and features into \(m\) clusters simultaneously:
\[
C_{r}\colon \{\mathbf{x}_1, \dots, \mathbf{x}_n\} 
\;\to\; \{\hat{\mathbf{x}}_1, \dots, \hat{\mathbf{x}}_g\}, 
\quad
C_{c}\colon \{\mathbf{y}_1, \dots, \mathbf{y}_d\} 
\;\to\; \{\hat{\mathbf{y}}_1, \dots, \hat{\mathbf{y}}_m\}.
\]
Here, \(C_r\) assigns each instance \(\mathbf{x}_i\) to one of the rows \(g\) and \(C_c\) assigns each characteristic \(\mathbf{y}_j\) to one of the columns \(m\). Reordering rows and columns according to these maps yields a block structure called \emph{co-clusters}.

\begin{figure}[htbp]
    \centering
    \begin{subfigure}[b]{0.45\textwidth}
        \includegraphics[width=\textwidth]{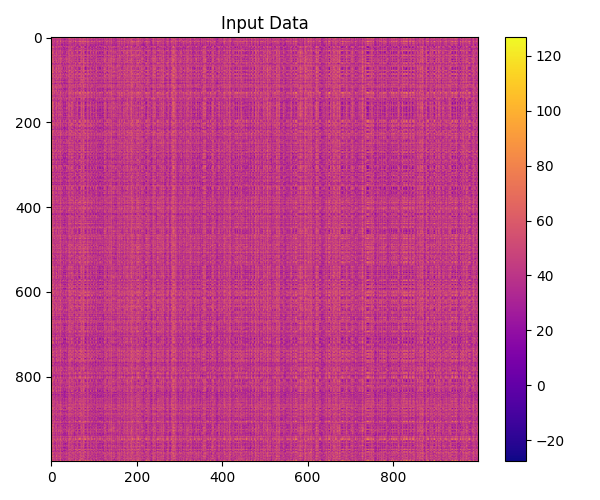} 
        \caption{Data Input}
        \label{fig:sub1}
    \end{subfigure}
    \hfill
    \begin{subfigure}[b]{0.45\textwidth}
        \includegraphics[width=\textwidth]{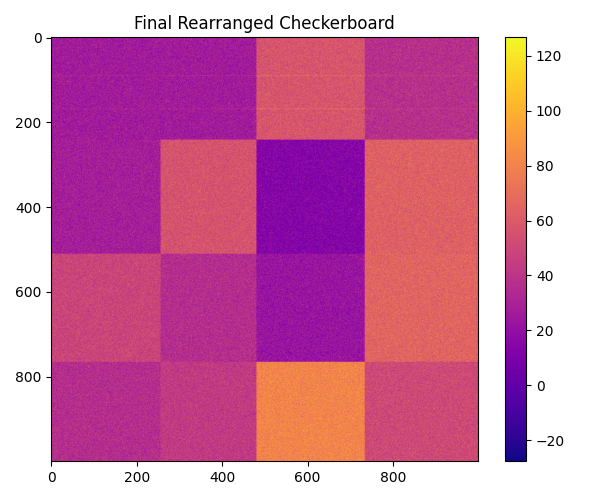} 
        \caption{Final Co-clusters}
        \label{fig:sub2}
    \end{subfigure}

    \caption{Plots showing the Inital input data matrix and final rearranged Co-cluster Checkerboard on Synthetic noisy data.}
    \label{fig:two_side_by_sidea}
\end{figure}

\subsection*{Discrete Variables and Their Partitions}

\begin{itemize}
  \item Let \(X_r\) be a discrete random variable taking values in \(\{\mathbf{x}_1, \dots, \mathbf{x}_n\}\).
  \item Let \(X_c\) be a discrete random variable taking values in \(\{\mathbf{y}_1, \dots, \mathbf{y}_d\}\).
  \item The joint distribution on the original data is \(p(X_r, X_c)\).
\end{itemize}

We define two new random variables \(\hat{X}_r = C_r(X_r)\) and \(\hat{X}_c = C_c(X_c)\) taking values in
\[
\hat{X}_r \in \{\hat{\mathbf{x}}_1, \dots, \hat{\mathbf{x}}_g\},
\quad
\hat{X}_c \in \{\hat{\mathbf{y}}_1, \dots, \hat{\mathbf{y}}_m\}.
\]
The induced distribution \(p(\hat{X}_r, \hat{X}_c)\) characterizes the co-clusters.

\begin{algorithm}[htbp]
\caption{Scalable Robust Bayesian Co-Clustering with Compositional ELBOs}
\label{alg:svdlcc_joint_concise}
\begin{algorithmic}[1]
\Require 
  $\mathbf{X}\in\mathbb{R}^{n\times d}$,
  row/column cluster counts $(g,m)$,
  hyperparameters $(\lambda_2,\lambda_3,\lambda_4,\lambda_5)$,
  row/col scale vectors $(\mathbf{f}^{(r)},\mathbf{f}^{(c)})$,
  priors $p_{\theta_r}(\mathbf{z}),\,p_{\theta_c}(\mathbf{z}),\,p_{\theta_{rc}}(\mathbf{z}_{rc})$,
  \(\text{max\_epochs}\).

\State Initialize row/column encoders/decoders $(\theta_r^e,\theta_r^d,\theta_c^e,\theta_c^d)$
\State Initialize joint encoder/decoder $(\theta_{rc}^e,\theta_{rc}^d)$
\State Initialize all mixture parameters $\{\pi_r^c,\mu_r^c,\sigma_r^c\}$, $\{\pi_c^k,\mu_c^k,\sigma_c^k\}$, $\{\pi_{rc}^m,\mu_{rc}^m,\Sigma_{rc}^m\}$

\For{epoch = 1 to \(\text{max\_epochs}\)}
  \State \textbf{ Update Scale Vectors:} 
         $\mathbf{f}^{(r)} \gets \text{RowScale}(\theta_r^e)$, 
         $\mathbf{f}^{(c)} \gets \text{ColScale}(\theta_c^e)$
  \Statex

  \State $J_{\mathrm{row}} \gets 0$
  \For{\textbf{mini-batch of row vectors} $\{\mathbf{x}_i\}$}
    \State $(\mu'_{x_i}, \sigma_{x_i}) \gets f_r(\mathbf{x}_i; \theta_r^e)$
    \State $\hat{\mu}_{x_i} \gets \mathbf{f}^{(r)} \odot \mu'_{x_i}$ \Comment{scaled means}
    \State $\mathbf{z}_i \sim \mathcal{N}(\hat{\mu}_{x_i}, \sigma_{x_i}^2)$
    \State $J_{\mathrm{row}} \mathrel{+}= \Bigl(-\log p_{\theta_r^d}(\mathbf{x}_i\mid\mathbf{z}_i)\Bigr)
                     \;+\;
                     D_{\mathrm{KL}}\bigl(\mathcal{N}(\mu'_{x_i},\sigma_{x_i}^2)\,\|\,p_{\theta_r}\bigr)$
  \EndFor
  \State $J_{\mathrm{row}} \gets \lambda_3\,J_{\mathrm{row}} \;+\;\lambda_2\,\|\theta_r\|^2$

  \Statex
  \State $J_{\mathrm{col}} \gets 0$
  \For{\textbf{mini-batch of column vectors} $\{\mathbf{y}_j\}$}
    \State $(\mu'_{y_j},\sigma_{y_j}) \gets f_c(\mathbf{y}_j;\theta_c^e)$
    \State $\hat{\mu}_{y_j} \gets \mathbf{f}^{(c)} \odot \mu'_{y_j}$ 
    \State $\mathbf{z}_j \sim \mathcal{N}(\hat{\mu}_{y_j},\sigma_{y_j}^2)$
    \State $J_{\mathrm{col}} \mathrel{+}= \Bigl(-\log p_{\theta_c^d}(\mathbf{y}_j\mid\mathbf{z}_j)\Bigr)
                     \;+\;
                     D_{\mathrm{KL}}\bigl(\mathcal{N}(\mu'_{y_j},\sigma_{y_j}^2)\,\|\,p_{\theta_c}\bigr)$
  \EndFor
  \State $J_{\mathrm{col}} \gets \lambda_3\,J_{\mathrm{col}} \;+\;\lambda_2\,\|\theta_c\|^2$

  \Statex
  \State $J_{\mathrm{joint}} \gets 0$
  \For{\textbf{mini-batch of cell pairs} $(i,j)$}
    \State $\mathbf{z}_i \gets f_r(\mathbf{x}_i;\theta_r^e)$,\,
           $\mathbf{z}_j \gets f_c(\mathbf{y}_j;\theta_c^e)$
    \State $(\mu_{rc}, \Sigma_{rc}) \gets f_{rc}(\mathbf{z}_i,\mathbf{z}_j;\theta_{rc}^e)$
    \State $\mathbf{z}_{rc}\sim\mathcal{N}(\mu_{rc},\Sigma_{rc})$ \Comment{or GMM sampling}
    \State $J_{\mathrm{joint}} \mathrel{+}= \Bigl(-\log p_{\theta_{rc}^d}(X_{i,j}\mid\mathbf{z}_{rc}, z_{r},z_{c}\Bigr)
                        \;+\;
                        D_{\mathrm{KL}}\!\Bigl(q_{\theta_{rc}^e}\,\|\,p_{\theta_{rc}}\Bigr)$
  \EndFor
  \State $J_{\mathrm{joint}} \gets \lambda_5\,J_{\mathrm{joint}}$

  \Statex
  \State Compute row cluster posteriors $\{\gamma_{r(i)}^c\}$, 
         column cluster posteriors $\{\gamma_{c(j)}^k\}$
  \State $I(\hat{X};\hat{Y}) \gets \text{MI}\bigl(\{\gamma_{r(i)}^c\},\{\gamma_{c(j)}^k\}\bigr)$
  \State $J_{\mathrm{MI}} \gets \lambda_4\Bigl(1 - \frac{I(\hat{X};\hat{Y})}{I(X;Y)}\Bigr)$

  \Statex
  \State $J_{\mathrm{total}}
     \gets
       J_{\mathrm{row}}
       + J_{\mathrm{col}}
       + J_{\mathrm{joint}}
       + J_{\mathrm{MI}}$
  \State Update all parameters 
         $\{\theta_r,\theta_c,\theta_{rc},\pi_r,\pi_c,\pi_{rc},\dots\}$
         via backprop w.r.t.\ $J_{\mathrm{total}}$

\EndFor

\Statex
\State \textbf{Return:} 
  row clusters $\{\gamma_{r(i)}^c\}$,\;
  column clusters $\{\gamma_{c(j)}^k\},\;$
  \text{joint cluster memberships }$ \bigl\{\gamma_{rc}(m\mid i,j)\bigr\}$.

\end{algorithmic}
\end{algorithm}

\section{Scaled Variational Co-Clustering}

In our \textit{Scaled Variational Co-Clustering} approach, we employ two separate Variational Autoencoder (VAE) models- one for instances (rows) and one for features (columns) each equipped with a Gaussian mixture model (GMM) prior to enforce cluster structure within the latent space. The loss formulation for these were derived from \cite{vade} On the instance side, we learn an encoder--decoder pair \((f_r, g_r)\) along with GMM parameters \(\{\pi_r^c, \mu_r^c, \sigma_r^c\}\), optimizing a negative Evidence Lower Bound (negative-ELBO) loss that combines reconstruction fidelity with a KL term to align the encoder's posterior distribution to the mixture-of-Gaussians prior. This gives us the soft clustering assignments \(\{\gamma_{r(i)}^c\}\) representing how each instance \(i\) belongs to each row-cluster component \(c\). Similarly on the feature side, with encoder-decoder \((f_c, g_c)\) and mixture parameters \(\{\pi_c^k, \mu_c^k, \sigma_c^k\}\), produces column-cluster assignments \(\{\gamma_{c(j)}^k\}\). To enforce that the row and column cluster assignments jointly capture the data's underlying co-structure, we incorporate a mutual information based cross-loss that pushes the co-clustered distribution \(p(\hat{X},\hat{Y})\) to reflect as much of the original mutual information \(I(X; Y)\) as possible. By combining these three components VAE-based reconstruction, GMM-regularized latent representations, and mutual information maximization we achieve a fully generative, robust, and scalable framework that discovers coherent clusters along both instance and feature dimensions.

\begin{figure*}[htbp]
  \centering
  \includegraphics[width=\textwidth]{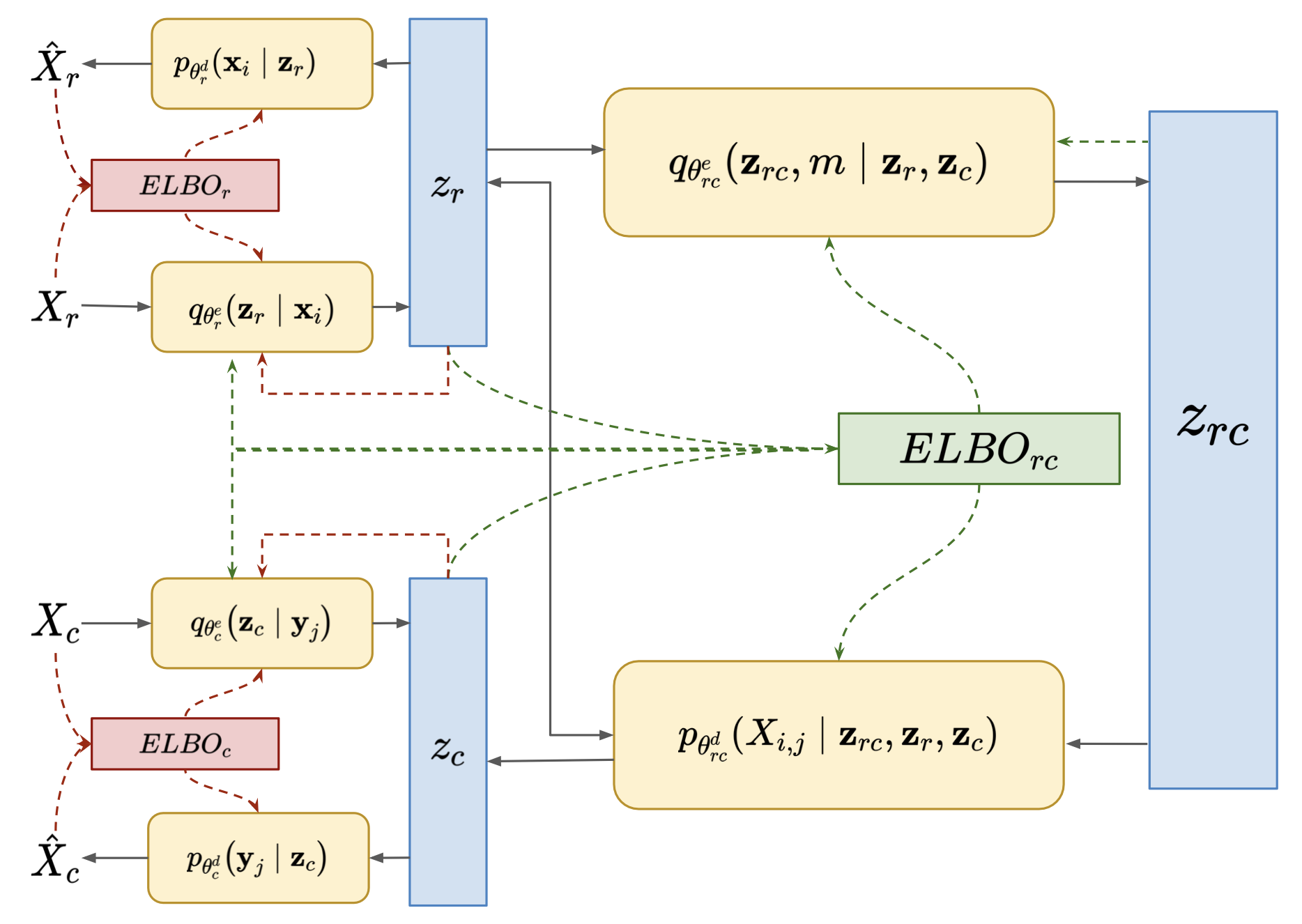}
  \caption{Variational Latent Co Clustering architecture.}
  \label{fig:arch}
\end{figure*}

\subsection{Prior Selection}
In our co-clustering framework, the choice of prior plays a critical role in shaping the latent space and enforcing clustering behavior. One option is to use a \emph{nonparametric} prior, such as the Dirichlet Process (DP), which is commonly employed in models like Dirichlet Process Mixture Models. While DP priors provide a flexible and adaptive solution for clustering, inference often becomes computationally expensive for the large datasets typically encountered in co-clustering. Moreover, the nonparametric nature of DP priors can lead to overfitting in high-dimensional latent spaces.

Another potential prior is the \emph{Gaussian Process (GP)}. Although GPs model correlations between data points via a kernel and can capture smoothness or manifold structures in the latent space, they do not inherently provide a discrete clustering mechanism. As a result, a post-training clustering step (e.g., spectral clustering) would be required, which disrupts the end-to-end nature of the co-clustering framework. Additionally, GPs incur $O(N^3)$ computational complexity due to the need to invert an $N \times N$ covariance matrix, making them infeasible for the larger datasets commonly encountered in co-clustering tasks. Although sparse GP methods can reduce computational overhead, they still do not provide a discrete clustering mechanism because they rely on a single continuous distribution.

For these reasons, we instead use a prior \emph{Gaussian Mixture Model (GMM)}, which naturally induces clustering through mixture components and provides soft assignments of data points to clusters. GMMs scale linearly with both the number of clusters and the number of data points, making them more computationally efficient for large datasets. Furthermore, the mixture-of-Gaussians structure encourages well-defined cluster geometry in the latent embeddings, which helps regularize against noise and improves interpretability. Finally, we derive the cluster assignments directly from the GMM posterior to compute our mutual information loss, allowing us to optimize the co-clustering objective in an end-to-end fashion.
.
\subsection{Instance-Side Loss}

We have a variational row autoencoder,  \textbf{Encoder} \(f_r(\mathbf{x}_i; \theta_r^e)\) outputs parameters of the approximate posterior, e.g.\ 
    \(\tilde{\boldsymbol{\mu}}_i,\, \tilde{\boldsymbol{\sigma}}_i\)., \textbf{Decoder} \(g_r(\mathbf{z};\theta_r^d)\) reconstructs \(\mathbf{x}_i\) from a latent \(\mathbf{z}\).

Instead of a single Gaussian prior, we define:
\[
p(\mathbf{z})
~\;=\;
\sum_{c=1}^{g} \pi_r^c\,\mathcal{N}\bigl(\mathbf{z}; \mu_r^c,\sigma_r^c\bigr).
\]
Here \(c \in \{1,\dots,g\}\) indexes the row-cluster components. The parameters are:
\(\pi_r^c\) (the cluster mixing weights), and \(\mu_r^c, \sigma_r^c\) (the mean and std of each mixture component in latent space).
\noindent

The negative Evidence Lower BOund (negative-ELBO) on the row side derived from \cite{vade}is:

\begin{align}
\mathcal{L}_{\mathrm{ELBO}}^{(\mathrm{row})}(\mathbf{x}_i)
&=\;
\mathbb{E}_{q(\mathbf{z},\,c \mid \mathbf{x}_i)}
\Bigl[
  \log p(\mathbf{x}_i \mid \mathbf{z})
  \;+\;
  \log p(\mathbf{z} \mid c)
  \nonumber\\
&\quad\;+\;
  \log p(c)
  \;-\;
  \log q(\mathbf{z}\mid \mathbf{x}_i)
  \;-\;
  \log q(c\mid \mathbf{x}_i)
\Bigr]\nonumber\\
&=\;
\underbrace{\mathbb{E}_{q(\mathbf{z}\mid \mathbf{x}_i)}
  \bigl[\log p(\mathbf{x}_i\mid \mathbf{z})\bigr]}_{\text{reconstruction term}}
\;-\;
D_{\mathrm{KL}}\Bigl(
  q(\mathbf{z},c\mid \mathbf{x}_i)
  \;\bigl\|\;
  p(\mathbf{z},c)
\Bigr).
\end{align}

\begin{equation}
\begin{aligned}
-\mathcal{L}_{\mathrm{ELBO}}^{(\mathrm{row})}
  (\mathbf{x}_i;\,\theta_r^e, \theta_r^d)
&=
-\,\mathbb{E}_{q_{\theta_r^e}(\mathbf{z}\mid \mathbf{x}_i)}
  \Bigl[
    \log p_{\theta_r^d}(\mathbf{x}_i \mid \mathbf{z})
  \Bigr]\\
&\quad
+\,D_{\mathrm{KL}}\Bigl(
   q_{\theta_r^e}(\mathbf{z}\mid \mathbf{x}_i)
   \,\big\|\,
   p(\mathbf{z})
\Bigr).
\end{aligned}
\end{equation}

\noindent
     The first term \(\mathbb{E}_{q(\mathbf{z}\mid \mathbf{x}_i)}[\log p(\mathbf{x}_i \mid \mathbf{z})]\) is the reconstruction term (log-likelihood under the decoder).
     The second term is the KL divergence between the encoder’s posterior \(q(\mathbf{z}\mid \mathbf{x}_i)\) and the mixture-of-Gaussians prior \(p(\mathbf{z})\).

\noindent

In the total row-side loss \(J_1\) We add a weight regularization \(\|\theta_r\|^2\).

\noindent

The row cluster assignment, for example, \(i\) is:
\[
\gamma_{r(i)}^c
~=~
q(c \mid \mathbf{x}_i)
~\approx~
\frac{\pi_r^c\,\mathcal{N}\bigl(\mathbf{z}_i;\,\mu_r^c,\sigma_r^c\bigr)}
{\sum_{c'}\pi_r^{c'}\,\mathcal{N}\bigl(\mathbf{z}_i;\,\mu_r^{c'},\sigma_r^{c'}\bigr)},
\]
where \(\mathbf{z}_i\) is sampled from \(q(\mathbf{z}\mid \mathbf{x}_i)\). The soft membership weights \(\{\gamma_{r(i)}^c\}\) indicate how the row \(i\) is distributed in the row groups.

\begin{align}
J_{1} &= 
\underbrace{\lambda_{1}\,\|\theta_{r}\|^{2}}_{\text{(regularizer)}}
~+~
\underbrace{\lambda_{2}\sum_{i=1}^n \Bigl(-\,\mathcal{L}_{\mathrm{ELBO}}^{(\mathrm{row})}(\mathbf{x}_i)\Bigr)}_{\text{VAE negative-ELBO on rows}}
\end{align}

Directly differentiating \(-\log p(\mathbf{z})\) w.r.t.\ the mixture parameters \(\theta_r\) in the KL term leads to high-variance score-function gradients. Hence we apply the \emph{doubly reparameterized} estimator. 
\newline
When using \(K\) importance samples \(\mathbf{z}_1,\dots,\mathbf{z}_K\sim q_{\theta_r^e}(\mathbf{z}\mid \mathbf{x}_i)\), define
\[
   w_k
   ~\;=\;
   \frac{
     p_{\theta_r}(\mathbf{z}_k)\;
     p_{\theta_r^d}(\mathbf{x}_i \mid \mathbf{z}_k)
   }{
     q_{\theta_r^e}(\mathbf{z}_k \mid \mathbf{x}_i)
   },
   \quad
   \widetilde{w}_k
   ~\;=\;
   \frac{w_k}{\sum_{j=1}^K w_j}.
\]
Here \(p_{\theta_r}(\mathbf{z}_k)\) is the prior mixture of Gaussians and \(q_{\theta_r^e}(\mathbf{z}_k \mid \mathbf{x}_i)\) is the encoder.

The gradient of the decoder (the expectation of \(-\log p_{\theta_r^d}(\mathbf{x}_i\mid \mathbf{z})\)) w.r.t.\ \(\theta_r^d\) is:
\[
   \nabla_{\theta_r^d}
   \Bigl[
     \mathbb{E}_{\,q_{\theta_r^e}}(\,-\log p_{\theta_r^d}(\mathbf{x}_i\mid \mathbf{z}))
   \Bigr]
   \;=\;
   \mathbb{E}_{\,\mathbf{z}\sim q_{\theta_r^e}}
   \Bigl[
     -\,\nabla_{\theta_r^d}\,\log p_{\theta_r^d}(\mathbf{x}_i\mid \mathbf{z})
   \Bigr].
\]
With multiple samples, this becomes:
\[
  \widehat{\nabla}_{\theta_r^d}
  =
  -\,\sum_{k=1}^K
    \widetilde{w}_k
    \;\nabla_{\theta_r^d}\,\log p_{\theta_r^d}(\mathbf{x}_i\mid \mathbf{z}_k),
\]
where each \(\mathbf{z}_k\) is drawn (reparameterized) from \(q_{\theta_r^e}\).

We address the  encoder score-function term
\(\nabla_{\theta_r^e}\log q_{\theta_r^e}(\mathbf{z}\mid \mathbf{x}_i)\).
Using the double reparameterization identity \cite{tucker2018doubly}, we obtain:
\[
   \widehat{\nabla}_{\theta_r^e}^{\mathrm{DREGs}}
   =
   \sum_{k=1}^K
   \widetilde{w}_k^{2}\;\nabla_{\mathbf{z}_k}\,\log w_k
   \;\nabla_{\theta_r^e}\,\mathbf{z}_k,
   \quad
   \mathbf{z}_k = T_{q}(\boldsymbol{\epsilon}_k;\,\theta_r^e).
\]
Here, \(\mathbf{z}_k\) is the usual reparameterized sample from the encoder distribution.

To handle \(\nabla_{\theta_r}\bigl[-\log p_{\theta_r}(\mathbf{z})\bigr]\) in the KL term, we use the generalized DREGs, which reparameterizes each \(\mathbf{z}_k\).  The resulting estimator is:
\[
  \widehat{\nabla}_{\theta_r}^{\mathrm{GDREGs}}
  =
  \sum_{k=1}^{K}
  \Bigl\{
     \underbrace{\widetilde{w}_k
     \,\nabla_{\mathbf{z}_k}\,\log p_{\theta_r^d}(\mathbf{x}_i\mid \mathbf{z}_k)}_{\text{recon.\ gradient part}}
     \;-\;
     \underbrace{\widetilde{w}_k^2
     \,\nabla_{\mathbf{z}_k}\,\log w_k}_{\text{mixture prior correction}}
  \Bigr\}
  \;\nabla_{\theta_r}\,T_{p}\!\bigl(\widetilde{\epsilon}_k;\,\theta_r\bigr)\Big|_{\widetilde{\epsilon}_k=\,T_{p}^{-1}(\mathbf{z}_k;\theta_r)},
\]
where \(\mathbf{z}_k\) is temporarily viewed as a sample from the prior reparameterization
\(\mathbf{z} = \mu_r^c + \sigma_r^c\,\boldsymbol{\epsilon}\) (after also sampling the discrete mixture component \(c \sim \mathrm{Cat}(\{\pi_r^c\})\)).

We now augment the loss with the contrastive loss derived from \cite{menon2022forget}, $c(\mathbf{x},\mathbf{z})$.  
For each mini-batch of size \(K\), we draw pairs \(\{(x_i, z_i)\}_{i=1}^K\), where 
\(z_i \sim q_{\theta_r^e}(z \mid x_i)\).  
Define a positive scalar function \(f_\psi(x,z)\) (e.g., $f_\psi = \exp[h_\psi(x,z)]$) and set:
\[
c(\mathbf{x}, \mathbf{z})
~=\;
\sum_{i=1}^K
\log
\biggl(
  \frac{
    f_{\psi}\bigl(x_i,\,z_i\bigr)
  }{
    \sum_{j=1}^K
    f_{\psi}\bigl(x_j,\,z_i\bigr)
  }
\biggr).
\]
Maximizing $c(\mathbf{x},\mathbf{z})$ (i.e.\ minimizing its negative) encourages each latent $z_i$ to be identifiable with its corresponding $x_i$ compared to mismatched pairs $(x_j,z_i)$, thereby raising the mutual information $I(x;z)$ and \emph{preventing} the posterior from collapsing to the prior.

The row side loss hence becomes,
\[
\boxed{
\begin{aligned}
J_{1}^{\text{(total)}}
&=\;
\underbrace{\lambda_{1}\,\|\theta_{r}\|^{2}}_{\text{(regularizer)}}
~+~
\underbrace{\lambda_{2}\,\sum_{i=1}^n 
\Bigl(
   -\,\mathcal{L}_{\mathrm{ELBO}}^{(\mathrm{row})}(\mathbf{x}_i)
\Bigr)}_{\text{VAE negative-ELBO on rows}}
~+~
\underbrace{\lambda_{3}\,\sum_{\text{batches}}
\Bigl(
   -\,c(\mathbf{x},\mathbf{z})
\Bigr)}_{\text{contrastive InfoNCE term}}
\end{aligned}
}
\]

\subsection{Feature-Side Loss}
We do the same thing for columns \(\mathbf{y}_j\). 
     Let \(\{\mathbf{y}_j\}_{j=1,\dots,d}\) be the set of column vectors.
     We again set up an autoencoder \((f_c, g_c)\) and a mixture-of-Gaussians prior 
    \(\{\pi_c^k, \mu_c^k,\sigma_c^k\}\) (with \(k=1,\dots,m\) indexing the column clusters).
Similarly, we have a  negative-ELBO for columns,
\(-\,\mathcal{L}_{\mathrm{ELBO}}^{(\mathrm{col})}(\mathbf{y}_j)\), plus  reconstruction and weight regularization. Thus the \emph{column-side} loss is:
\[
\begin{aligned}
\mathcal{L}_{\mathrm{ELBO}}^{(\mathrm{col})}(\mathbf{y}_j)
&=\;
\mathbb{E}_{\,q(\mathbf{z},\,k \mid \mathbf{y}_j)} \Bigl[
  \log p(\mathbf{y}_j \mid \mathbf{z})
  \;+\;
  \log p(\mathbf{z} \mid k)
  \\
&\quad\;+\;
  \log p(k)
  \;-\;
  \log q(\mathbf{z}\mid \mathbf{y}_j)
  \;-\;
  \log q(k\mid \mathbf{y}_j)
\Bigr]
\\[6pt]
&=
\underbrace{\mathbb{E}_{\,q(\mathbf{z}\mid \mathbf{y}_j)}\bigl[\log p(\mathbf{y}_j\mid \mathbf{z})\bigr]}_{\text{reconstruction term}}
\;-\;
D_{\mathrm{KL}}\Bigl(
  q(\mathbf{z},k\mid \mathbf{y}_j)
  \;\bigl\|\;
  p(\mathbf{z},k)
\Bigr),
\end{aligned}
\]
where 
\(
p(\mathbf{z},k)
~\;=\;
p(\mathbf{z}\mid k)\,p(k).
\)
\begin{equation}
\begin{aligned}
-\mathcal{L}_{\mathrm{ELBO}}^{(\mathrm{col})}(\mathbf{y}_j;\,\theta_c^e, \theta_c^d)
~=~&
-\,\mathbb{E}_{q_{\theta_c^e}(\mathbf{z} \mid \mathbf{y}_j)}
\Bigl[\,
  \log p_{\theta_c^d}(\mathbf{y}_j \mid \mathbf{z})
\Bigr]
\\
&~+~
D_{\mathrm{KL}}
\Bigl(
   q_{\theta_c^e}(\mathbf{z}\mid \mathbf{y}_j)
   ~\Big\|\,
   p(\mathbf{z})
\Bigr).
\end{aligned}
\end{equation}
After training, the column-cluster assignment for feature \(j\) is:
\[
\gamma_{c(j)}^k
=
q(k \mid \mathbf{y}_j)
=
\frac{\pi_c^k\,\mathcal{N}\bigl(\mathbf{z}_j;\,\mu_c^k,\sigma_c^k\bigr)}
     {\sum_{k'}\pi_c^{k'}\,\mathcal{N}\bigl(\mathbf{z}_j;\,\mu_c^{k'},\sigma_c^{k'}\bigr)},
\]
where \(\mathbf{z}_j\sim q(\mathbf{z}\mid \mathbf{y}_j)\) from the column-side encoder.
Hence the column-side loss is:

\begin{align}    
J_{2} &=\;
\underbrace{\lambda_{3}\,\|\theta_c\|^2}_{\text{(regularizer)}} 
\;+\;
\underbrace{\lambda_{4}\sum_{j=1}^d \Bigl(-\,\mathcal{L}_{\mathrm{ELBO}}^{(\mathrm{col})}(\mathbf{y}_j)\Bigr)}_{\text{(negative-ELBO on columns)}}
\end{align}

As with the row side, the negative expected log-likelihood w.r.t.\ \(\theta_c^d\) is a direct pathwise derivative. So the gradient of the decoder is:
\[
   \nabla_{\theta_c^d}
   \Bigl[
     \mathbb{E}_{\,q_{\theta_c^e}(\mathbf{z}\mid \mathbf{y}_j)}
        \bigl(-\log p_{\theta_c^d}(\mathbf{y}_j\mid \mathbf{z})\bigr)
   \Bigr]
   ~=~
   \mathbb{E}_{\,q_{\theta_c^e}}
   \Bigl[
     -\,\nabla_{\theta_c^d}\,\log p_{\theta_c^d}(\mathbf{y}_j\mid \mathbf{z})
   \Bigr].
\]
In multi-sample (IWAE) style with weights \(w_k\), this becomes:
\[
   \widehat{\nabla}_{\theta_c^d}
   ~=~
   -\,\sum_{k=1}^K
     \widetilde{w}_k
     \;\nabla_{\theta_c^d}\,\log p_{\theta_c^d}\!\bigl(\mathbf{y}_j\mid \mathbf{z}_k\bigr),
\]
where \(\widetilde{w}_k = w_k / \sum_{r=1}^K w_r\).

We eliminate the naive score-function \(\nabla_{\theta_c^e}\,\log q_{\theta_c^e}\!\bigl(\mathbf{z}\mid \mathbf{y}_j\bigr)\) using the DREG approach:
\[
  \widehat{\nabla}_{\theta_c^e}^{\mathrm{DREGs}}
  ~=~
  \sum_{k=1}^K
  \widetilde{w}_k^{2}
  \;\nabla_{\mathbf{z}_k}\,\log w_k
  \;\nabla_{\theta_c^e}\,\mathbf{z}_k,
\]
with \(\mathbf{z}_k\sim q_{\theta_c^e}(\mathbf{z}\mid \mathbf{y}_j)\). Here,
\[
  w_k
  \;=\;
  \frac{
     p_{\theta_c}(\mathbf{z}_k)\,p_{\theta_c^d}(\mathbf{y}_j\mid \mathbf{z}_k)
  }{
     q_{\theta_c^e}(\mathbf{z}_k\mid \mathbf{y}_j)
  },
  \quad
  \widetilde{w}_k = \frac{w_k}{\sum_{\ell=1}^K w_\ell}.
\]
On the prior gradients, \(\nabla_{\theta_c}\,\log p_{\theta_c}(\mathbf{z}_k)\), we \emph{reparameterize} each \(\mathbf{z}_k\) as if from the mixture-of-Gaussians \(p_{\theta_c}(\mathbf{z})\):
\[
  \widehat{\nabla}_{\theta_c}^{\mathrm{GDREGs}}
  ~=~
  \sum_{k=1}^{K}
  \Bigl\{
    \widetilde{w}_k\;\nabla_{\mathbf{z}_k}\,\log p_{\theta_c^d}(\mathbf{y}_j\mid \mathbf{z}_k)
    ~-\;
    \widetilde{w}_k^{2}\;\nabla_{\mathbf{z}_k}\,\log w_k
  \Bigr\}
  \;\nabla_{\theta_c}\,T_{p_c}\bigl(\widetilde{\epsilon}_k;\,\theta_c\bigr)
  \Big|_{\widetilde{\epsilon}_k=\,T_{p_c}^{-1}(\mathbf{z}_k;\,\theta_c)},
\]
where \(T_{p_c}\) is the mixture reparameterization for columns (sample discrete \(k\sim\pi_c^k\), then Gaussian \(\boldsymbol{\epsilon}\sim \mathcal{N}(0,I)\)).

similar to the contrastive loss introduced in the row side loss , we introduce a contrastive loss for the column side as well.

\[
\boxed{
\begin{aligned}
J_{2}^{\text{(total)}}
&=\;
\underbrace{\lambda_{4}\,\|\theta_c\|^2}_{\text{(regularizer)}}
\;+\;
\underbrace{\lambda_{5}
\sum_{j=1}^d 
\Bigl(
  -\,\mathcal{L}_{\mathrm{ELBO}}^{(\mathrm{col})}(\mathbf{y}_j)
\Bigr)}_{\text{negative-ELBO on columns}}
\;+\;
\underbrace{\lambda_{6}
\sum_{\text{batches}}
\Bigl(
  -\,c\bigl(\mathbf{y}, \mathbf{z}\bigr)
\Bigr)}_{\text{contrastive loss term}}
\end{aligned}
}
\]

\subsection{Joint Space}
In addition to the row‐side and column‐side VAEs, we introduce a third latent variable,
\(\mathbf{z}_{rc}\), for each cell \((i,j)\) in the data matrix. The goal is to capture
\emph{row--column interactions} that are fully represented by the separate row and
column embeddings alone. In many practical scenarios, individual cells can contain noise 
or anomalies that do not align neatly with a single row or column embedding. By 
introducing \(\mathbf{z}_{rc}\), we provide an additional stage that allows local deviations and noise to be modeled separately from the global row or 
column factors. This decomposition helps ensure that the row‐ and column‐side embeddings 
retain a cleaner, more interpretable structure, while \(\mathbf{z}_{rc}\) absorbs the 
cell‐specific sources of variation. \\

The Joint Encoder \(\;f_{rc}(\mathbf{z}_i,\mathbf{z}_j;\,\theta_{rc}^e)\) takes the row‐side sample \(\mathbf{z}_i\sim q_{\theta_r^e}(\mathbf{z}_i\mid \mathbf{x}_i)\) and the column‐side sample \(\mathbf{z}_j\sim q_{\theta_c^e}(\mathbf{z}_j\mid \mathbf{y}_j)\) and outputs the distribution \(q_{\theta_{rc}^e}\!\bigl(\mathbf{z}_{rc}\mid \mathbf{z}_i,\mathbf{z}_j\bigr)\); meanwhile, the Joint Decoder \(\;g_{rc}\bigl(\mathbf{z}_{rc};\,\theta_{rc}^d\bigr)\) reconstructs the cell entry \(X_{i,j}\) by computing \(\widehat{X}_{i,j} = g_{rc}\bigl(\mathbf{z}_{rc};\,\theta_{rc}^d\bigr)\). For the Prior on \(\mathbf{z}_{rc}\), We use a GMM prior like before to allow for multi‐modal or block‐structured embeddings. With \(M\) mixture components,
      \[
        p_{\theta_{rc}}\!\bigl(\mathbf{z}_{rc}\bigr)
        \;=\;
        \sum_{m=1}^{M}
        \pi_{rc}^{m}
        \,\mathcal{N}\!\bigl(\mathbf{z}_{rc};\,
                             \mu_{rc}^{m},\,
                             \Sigma_{rc}^{m}\bigr),
        \quad
        \sum_{m=1}^{M}\,\pi_{rc}^{m}=1.
      \]
      Where,
      \(\{\pi_{rc}^m,\mu_{rc}^m,\Sigma_{rc}^m\}_{m=1}^M\)
      are learnable parameters.\\
      
Let \(\mathbf{z}_i\sim q_{\theta_r^e}(\mathbf{z}_i\mid \mathbf{x}_i)\) and
\(\mathbf{z}_j\sim q_{\theta_c^e}(\mathbf{z}_j\mid \mathbf{y}_j)\).  For each cell
\((i,j)\), the \emph{joint‐side} negative ELBO is
 conditioned on \(\mathbf{z}_i,\mathbf{z}_j\). We introduce a discrete mixture index
\(m\in\{1,\dots,M\}\) such that
\(\;p(m)=\pi_{rc}^m\),
\(\;p\bigl(\mathbf{z}_{rc}\mid m\bigr)
=\mathcal{N}(\mathbf{z}_{rc};\mu_{rc}^m,\Sigma_{rc}^m)\),
we define the joint encoder
\(\;q_{\theta_{rc}^e}\!\bigl(\mathbf{z}_{rc},\,m\mid \mathbf{z}_i,\mathbf{z}_j\bigr)\).
The mixture ELBO can be written as:
\[
\begin{aligned}
\mathcal{L}_{\mathrm{ELBO}}^{(\mathrm{joint})}\!\bigl(X_{i,j}\bigr)
&=\;
\mathbb{E}_{\,q(\mathbf{z}_{rc},\,m\mid \mathbf{z}_i,\mathbf{z}_j)}
\Bigl[
  \log p_{\theta_{rc}^d}\!\bigl(X_{i,j}\mid \mathbf{z}_{rc}\bigr)
  ~+~
  \log p_{\theta_{rc}}\!\bigl(\mathbf{z}_{rc},\,m\bigr)
  ~-~
  \log q_{\theta_{rc}^e}\!\bigl(\mathbf{z}_{rc},\,m\mid \mathbf{z}_i,\mathbf{z}_j\bigr)
\Bigr].
\end{aligned}
\]
Hence, the negative ELBO is
\begin{equation}
\label{eq:joint-side-elbo}
-\mathcal{L}_{\mathrm{ELBO}}^{(\mathrm{joint})}\!\bigl(X_{i,j}\bigr)
\;=\;
-\,\mathbb{E}_{
   \,q(\mathbf{z}_{rc}\mid \mathbf{z}_i,\mathbf{z}_j)
 }
 \Bigl[
   \log p_{\theta_{rc}^d}\!\bigl(X_{i,j}\mid \mathbf{z}_{rc}\bigr)
 \Bigr]
~+~
D_{\mathrm{KL}}\Bigl(
  q(\mathbf{z}_{rc},\,m\mid \mathbf{z}_i,\mathbf{z}_j)
  \,\Big\|\,
  p_{\theta_{rc}}(\mathbf{z}_{rc},\,m)
\Bigr).
\end{equation}

As $z_rc$ take inputs sampled from the instance and feature encoders, we expand the joint elbo as:

\begin{equation}
\label{eq:joint-side-elbo-expanded}
\begin{aligned}
-\mathcal{L}_{\mathrm{ELBO}}^{(\mathrm{joint})}\!\bigl(X_{i,j}\bigr)
&=\;
-\,\mathbb{E}_{\substack{
   \mathbf{z}_r \sim q_{\theta_r^e}(\mathbf{z}_r \mid \mathbf{x}_i)\\
   \mathbf{z}_c \sim q_{\theta_c^e}(\mathbf{z}_c \mid \mathbf{y}_j)\\
   \mathbf{z}_{rc},m \,\sim\,q_{\theta_{rc}^e}(\mathbf{z}_{rc},m\mid \mathbf{z}_r,\mathbf{z}_c)
}}
\Bigl[
  \log p_{\theta_{rc}^d}\!\bigl(X_{i,j}\mid \mathbf{z}_{rc},\mathbf{z}_r,\mathbf{z}_c\bigr)
\Bigr]
\\
&\quad
+\;
\mathbb{E}_{\mathbf{z}_r \sim q_{\theta_r^e},\;\mathbf{z}_c \sim q_{\theta_c^e}}
\Bigl[
  D_{\mathrm{KL}}\Bigl(
    q_{\theta_{rc}^e}(\mathbf{z}_{rc},m \mid \mathbf{z}_r,\mathbf{z}_c)
    \;\Big\|\;
    p_{\theta_{rc}}(\mathbf{z}_{rc},m)
  \Bigr)
\Bigr].
\end{aligned}
\end{equation}

Summing over all \((i,j)\) yields the total joint‐side loss:
\begin{equation}
\label{eq:joint-loss}
J_{\mathrm{joint}}
\;=\;
\lambda_{5}\,
\sum_{i=1}^n
\sum_{j=1}^d
\Bigl[
  -\,\mathcal{L}_{\mathrm{ELBO}}^{(\mathrm{joint})}\bigl(X_{i,j}\bigr)
\Bigr],
\end{equation} \\
The joint‐side mixture provides \emph{cell‐level cluster assignments}. Specifically, given a learned mixture prior
\[
  p_{\theta_{rc}}\!\bigl(\mathbf{z}_{rc},m\bigr)
  \;=\;
  \pi_{rc}^m
  \,\mathcal{N}\!\bigl(\mathbf{z}_{rc};\mu_{rc}^m,\Sigma_{rc}^m\bigr),
\]
we  define the posterior probability that cell \((i,j)\) belongs to the
\(m\)-th \emph{joint cluster}:
\[
  \gamma_{rc}(m\mid i,j)
  \;=\;
  q\bigl(m\mid \mathbf{z}_{rc},\mathbf{z}_i,\mathbf{z}_j\bigr)
  \;\approx\;
  \frac{
    \pi_{rc}^m\,
    \mathcal{N}\!\bigl(\mathbf{z}_{rc};\,\mu_{rc}^m,\Sigma_{rc}^m\bigr)
  }{
    \sum_{m'=1}^M
      \pi_{rc}^{m'}\,
      \mathcal{N}\!\bigl(\mathbf{z}_{rc};\,\mu_{rc}^{m'},\Sigma_{rc}^{m'}\bigr)
  }.
\]
Hence, each cell~\((i,j)\) obtains a \emph{soft} membership distribution across the
\(\{1,\dots,M\}\) components from which we take the \(\arg\max\) to yield a single
\emph{joint block} label \(m\). This allows the model to discover and interpret
\(\mathbf{z}_{rc}\) as capturing per‐cell or ``block'' clusters, in addition to the
row‐side and column‐side cluster memberships.
\\
The decoder
$p_{\theta_{rc}^d}\!\bigl(X_{i,j}\mid \mathbf{z}_{rc},\mathbf{z}_r,\mathbf{z}_c\bigr)$
depends on all three latents: $\mathbf{z}_r$, $\mathbf{z}_c$, and $\mathbf{z}_{rc}$. Let $(\mathbf{z}_{r,k}, \mathbf{z}_{c,k})$ be $K$ reparameterized samples from
\[
  q_{\theta_r^e}(\mathbf{z}_r \mid \mathbf{x}_i)
  \;\;\;\text{and}\;\;\;
  q_{\theta_c^e}(\mathbf{z}_c \mid \mathbf{y}_j),
\]
respectively.
Conditioned on $(\mathbf{z}_{r,k}, \mathbf{z}_{c,k})$, draw the mixture index $m_k \sim \mathrm{Cat}(\{\pi_{rc}^m\})$ and the cell‐level latent
$\mathbf{z}_{rc,k} = \mu_{rc}^{\,m_k} + \Sigma_{rc}^{\,m_k}\,\boldsymbol{\epsilon}_k$,
via
$q_{\theta_{rc}^e}(\mathbf{z}_{rc},m\mid \mathbf{z}_{r,k},\mathbf{z}_{c,k})$.
Define the importance weight:
\begin{equation}
\label{eq:joint_wk}
w_k
~=\;
\frac{
  p_{\theta_{rc}}\!\bigl(\mathbf{z}_{rc,k},\,m_k\bigr)
  \;\times\;
  p_{\theta_{rc}^d}\!\bigl(X_{i,j}\,\big|\,
      \mathbf{z}_{rc,k},\,\mathbf{z}_{r,k},\,\mathbf{z}_{c,k}\bigr)
}{
  q_{\theta_{rc}^e}\!\bigl(\mathbf{z}_{rc,k},\,m_k\,\big|\,
    \mathbf{z}_{r,k},\,\mathbf{z}_{c,k}\bigr)
},
\quad
\widetilde{w}_k
~=\;
\frac{\,w_k\,}{\sum_{\ell=1}^K w_{\ell}}.
\end{equation}

For the decoder, The gradient w.r.t.\ the \emph{negative} log‐likelihood term is the standard IWAE‐style
\emph{pathwise} derivative:
\[
\nabla_{\theta_{rc}^d}
\Bigl[
  -\,\mathbb{E}_{\,q_{\theta_r^e},\,q_{\theta_c^e},\,q_{\theta_{rc}^e}}
     \log p_{\theta_{rc}^d}\!\bigl(X_{i,j}
       \mid \mathbf{z}_{rc},\,\mathbf{z}_r,\,\mathbf{z}_c\bigr)
\Bigr]
~\approx~
-\,\sum_{k=1}^K
  \widetilde{w}_k
  \;\nabla_{\theta_{rc}^d}
  \log p_{\theta_{rc}^d}\!\bigl(X_{i,j}
    \mid \mathbf{z}_{rc,k},\,\mathbf{z}_{r,k},\,\mathbf{z}_{c,k}\bigr).
\]
Because $(\mathbf{z}_{rc,k},\,\mathbf{z}_{r,k},\,\mathbf{z}_{c,k})$ are reparameterized
samples, this derivative remains purely pathwise.

For the encoder, We would  score‐function terms
$\nabla_{\theta_{rc}^e}\log q_{\theta_{rc}^e}$.
The \emph{Doubly Reparameterized Gradient} (DREG) estimator
\cite{tucker2018doubly} transforms these into lower‐variance pathwise derivatives. So for each $k$,
\[
\nabla_{\theta_{rc}^e}^{(\mathrm{DREG})}
~\approx~
\sum_{k=1}^K
  \widetilde{w}_k^{\,2}
  \;\nabla_{\mathbf{z}_{rc,k}}\,
  \log \bigl(
    w_k
  \bigr)
  \;\nabla_{\theta_{rc}^e}\,\mathbf{z}_{rc,k},
\]
where $\mathbf{z}_{rc,k}$ (and $m_k$) are reparameterized from
$q_{\theta_{rc}^e}$.  Since
\[
\log\bigl(w_k\bigr)
~=~
\log p_{\theta_{rc}}\!\bigl(\mathbf{z}_{rc,k},m_k\bigr)
~+~
\log p_{\theta_{rc}^d}\!\bigl(X_{i,j}\mid
      \mathbf{z}_{rc,k},\,\mathbf{z}_{r,k},\,\mathbf{z}_{c,k}\bigr)
~-\;
\log q_{\theta_{rc}^e}\!\bigl(\mathbf{z}_{rc,k},\,m_k\mid
      \mathbf{z}_{r,k},\,\mathbf{z}_{c,k}\bigr),
\]
we take pathwise derivatives w.r.t.\ $\mathbf{z}_{rc,k}$.
\\
Because the decoder depends on $(\mathbf{z}_r,\mathbf{z}_c)$ \emph{as well} as
$\mathbf{z}_{rc}$, gradients also flow through the row/column encoders:
\[
\nabla_{\theta_r^e},\;\nabla_{\theta_c^e}.
\]
Each $\mathbf{z}_{r,k}$ and $\mathbf{z}_{c,k}$ is a reparameterized sample from
$q_{\theta_r^e}(\mathbf{z}_r \mid \mathbf{x}_i)$ and
$q_{\theta_c^e}(\mathbf{z}_c \mid \mathbf{y}_j)$, respectively. Hence, we get pathwise
terms whenever the decoder
$\log p_{\theta_{rc}^d}\bigl(X_{i,j}\mid \mathbf{z}_{rc,k}, \mathbf{z}_{r,k}, \mathbf{z}_{c,k}\bigr)$
and/or the encoder
$\log q_{\theta_{rc}^e}\bigl(\mathbf{z}_{rc,k},m_k\mid \mathbf{z}_{r,k},\mathbf{z}_{c,k}\bigr)$
depend on $\mathbf{z}_{r,k}, \mathbf{z}_{c,k}$. By the chain rule,
\[
\nabla_{\theta_r^e} 
~
\approx~
\sum_{k=1}^K
  \widetilde{w}_k
  \;\nabla_{\mathbf{z}_{r,k}}\,
     \log p_{\theta_{rc}^d}\!\bigl(X_{i,j}\mid 
     \mathbf{z}_{rc,k},\mathbf{z}_{r,k},\mathbf{z}_{c,k}\bigr)
  \;\nabla_{\theta_r^e}\,\mathbf{z}_{r,k}
~+~
\sum_{k=1}^K
  \widetilde{w}_k^2
  \;\nabla_{\mathbf{z}_{r,k}}\,
  \log\!\Bigl[\tfrac{w_k}{\sum_{\ell=1}^K w_{\ell}}\Bigr]
  \;\nabla_{\theta_r^e}\,\mathbf{z}_{r,k},
\]
and similarly for $\nabla_{\theta_c^e}$.

Finally, the parameters of the mixture prior
$p_{\theta_{rc}}(\mathbf{z}_{rc},m)$ appear inside
$\log p_{\theta_{rc}}\!\bigl(\mathbf{z}_{rc,k},\,m_k\bigr)$, so we apply the
\emph{generalized DREG} approach \cite{tucker2018doubly}:
\[
\nabla_{\theta_{rc}}
~\approx~
\sum_{k=1}^K
\Bigl\{
   \widetilde{w}_k
   \,\nabla_{\mathbf{z}_{rc,k}}\,
   \log p_{\theta_{rc}^d}\!\bigl(X_{i,j}\mid 
     \mathbf{z}_{rc,k},\mathbf{z}_{r,k},\mathbf{z}_{c,k}\bigr)
   ~-~
   \widetilde{w}_k^{\,2}
   \,\nabla_{\mathbf{z}_{rc,k}}\,
   \log w_k
\Bigr\}
\;\nabla_{\theta_{rc}}\!\mathbf{z}_{rc,k}.
\]
Here, $\nabla_{\theta_{rc}}\!\mathbf{z}_{rc,k}$ follows from reparameterizing
$\mathbf{z}_{rc,k}$ using the mixture means $\mu_{rc}^m$ and covariances $\Sigma_{rc}^m$.

We now append a contrastive term 
\(\,c_{\mathrm{joint}}(X_{i,j},\,\mathbf{z}_{rc})\)
to capture cell-level dependence more robustly. For each mini-batch of cells 
\((i,j)\), we sample a “positive” pair 
\(\bigl(X_{i,j}, \mathbf{z}_{rc}\bigr)\) 
(where 
\(\mathbf{z}_{rc} \sim q_{\theta_{rc}^e}\!\bigl(\mathbf{z}_{rc}\mid \mathbf{z}_r,\mathbf{z}_c\bigr)\))
and create “negative” pairs by mismatching \(\mathbf{z}_{rc}\) with another cell’s observation.

We define the \emph{negative} of the contrastive objective as 
\(-\,c_{\mathrm{joint}}(X_{i,j},\,\mathbf{z}_{rc})\), 
so \emph{minimizing} it is equivalent to \emph{maximizing} the contrast.
We weight it by a scalar \(\lambda_{\mathrm{joint}}>0\) and obtain
\begin{equation}
\label{eq:joint-contrast-full}
\boxed{
J_{\mathrm{joint}}^{(\mathrm{aug})}
~=~
\underbrace{
  \lambda_{7}
  \sum_{i=1}^n
  \sum_{j=1}^d
  \Bigl[
    -\,\mathcal{L}_{\mathrm{ELBO}}^{(\mathrm{joint})}\bigl(X_{i,j}\bigr)
  \Bigr]
}_{\text{Negative-ELBO from \eqref{eq:joint-loss}}}
~+~
\underbrace{
  \lambda_{8}
  \sum_{\text{batches}}
  \Bigl[
    -\,c_{\mathrm{joint}}\!\bigl(X_{i,j},\,\mathbf{z}_{rc}\bigr)
  \Bigr]
}_{\text{contrastive regularizer}}
}.
\end{equation}

\noindent
Minimizing \eqref{eq:joint-contrast-full} will fit the joint-side mixture-of-Gaussians VAE via the standard negative-ELBO and maximize $c_{\mathrm{joint}}(\cdot)$, thereby preserving mutual information between each cell $X_{i,j}$ and its latent $\mathbf{z}_{rc}$.
\subsection{Cross-Loss (Mutual Information)}

We define a mutual information loss to enforce coherent row–column partitions and preserve the dependence between instances $X$ and features $Y$ under the learned co-cluster assignments. Let
\[
I(X;Y)
~=~
\sum_{\mathbf{x}_i} \sum_{\mathbf{y}_j}
  p(\mathbf{x}_i, \mathbf{y}_j)
  \;\log\!
  \Bigl[
    \tfrac{
      p(\mathbf{x}_i, \mathbf{y}_j)
    }{
      p(\mathbf{x}_i)\,p(\mathbf{y}_j)
    }
  \Bigr]
\]
be the Mutual information in the original data distribution $p(X,Y)$. 

Each row $i$ has a \emph{soft} membership distribution
\[
\gamma_{r(i)}
~=~
\bigl(\,\gamma_{r(i)}^{\,1},\dots,\gamma_{r(i)}^{\,g}\bigr)
\quad
\text{where}
\quad
\sum_{s=1}^g \gamma_{r(i)}^{\,s} \;=\;1,
\]
and each column $j$ has a \emph{soft} membership distribution
\[
\gamma_{c(j)}
~=~
\bigl(\,\gamma_{c(j)}^{\,1},\dots,\gamma_{c(j)}^{\,m}\bigr)
\quad
\text{where}
\quad
\sum_{t=1}^m \gamma_{c(j)}^{\,t} \;=\;1.
\]
We use these distributions to define an induced “co-cluster” random variable $(\hat{X}, \hat{Y})$, where
\[
\hat{X} \in \{1,\dots,g\},
\quad
\hat{Y} \in \{1,\dots,m\}.
\]
Specifically, the joint distribution $p(\hat{X}, \hat{Y})$ is obtained by marginalizing over \emph{all} rows and columns:
\[
p\!\bigl(\hat{X} = s,\, \hat{Y} = t\bigr)
~=\;
\sum_{i=1}^n \sum_{j=1}^d
  p(i,j)
  \;\gamma_{r(i)}^{\,s}
  \;\gamma_{c(j)}^{\,t},
\]
where $p(i,j) = \tfrac{1}{n\,d}$ if rows and columns are assumed uniformly likely. The induced marginals follow from summing out one index, e.g.
\[
p(\hat{X} = s)
~=~
\sum_{i=1}^n
  p(i)\,\gamma_{r(i)}^{\,s}
\quad\text{with}\quad
p(i) = \tfrac{1}{n},
\]
and similarly for $p(\hat{Y} = t)$.

The mutual information of the soft labels $(\hat{X},\hat{Y})$ is:
\[
I(\hat{X}; \hat{Y})
~=~
\sum_{s=1}^g
\sum_{t=1}^m
  p\!\bigl(\hat{X} = s,\,\hat{Y} = t\bigr)
  \,\log\!
  \Bigl[
    \tfrac{
      p(\hat{X}\!=s,\,\hat{Y}\!=t)
    }{
      p(\hat{X}\!=s)\,\;p(\hat{Y}\!=t)
    }
  \Bigr].
\]
\begin{equation}
J_3=\lambda_9\left(1-\frac{I(\hat{X} ; \hat{Y})}{I(X ; Y)}\right)=\lambda_9\left(1-\frac{\sum_{s, t} p\left(\hat{\mathbf{x}}_s, \hat{\mathbf{y}}_t\right) \log [\ldots]}{\sum_{i, j} p\left(\mathbf{x}_i, \mathbf{y}_j\right) \log [\ldots]}\right)
\end{equation}

\subsection*{Combined Objective}

\begin{equation}
\label{eq:combined-obj-contrastive}
\boxed{
\begin{aligned}
J_{\mathrm{total}}
~=\;&
\underbrace{
  \lambda_{1}\,\|\theta_r\|^{2}
  ~+~
  \lambda_{2}\,\sum_{i=1}^n
    \Bigl[-\,\mathcal{L}_{\mathrm{ELBO}}^{(\mathrm{row})}(\mathbf{x}_i)\Bigr]
  ~+~
  \lambda_{3}\,\sum_{\text{batches}}
    \Bigl[-\,c\bigl(\mathbf{x},\mathbf{z}\bigr)\Bigr]
}_{\text{row side (\,Instance-Side Loss)}}
\\[0.6em]
&\quad
+\;
\underbrace{
  \lambda_{4}\,\|\theta_c\|^{2}
  ~+~
  \lambda_{5}\,\sum_{j=1}^d
    \Bigl[-\,\mathcal{L}_{\mathrm{ELBO}}^{(\mathrm{col})}(\mathbf{y}_j)\Bigr]
  ~+~
  \lambda_{6}\,\sum_{\text{batches}}
    \Bigl[-\,c\bigl(\mathbf{y},\mathbf{z}\bigr)\Bigr]
}_{\text{column side (\,Feature-Side Loss)}}
\\[0.6em]
&\quad
+\;
\underbrace{
  \lambda_{7}\,\sum_{i=1}^n \sum_{j=1}^d
    \Bigl[-\,\mathcal{L}_{\mathrm{ELBO}}^{(\mathrm{joint})}\bigl(X_{i,j}\bigr)\Bigr]
  ~+~
  \lambda_{8}\,\sum_{\text{batches}}
    \Bigl[-\,c_{\mathrm{joint}}\!\bigl(X_{i,j},\,\mathbf{z}_{rc}\bigr)\Bigr]
}_{\text{joint side (\,Joint Space)}}
\\[0.6em]
&\quad
+\;
\underbrace{
  \lambda_{9}\,\Bigl(
    1 \;-\; \tfrac{I(\hat{X};\hat{Y})}{I(X;Y)}
  \Bigr)
}_{\text{cross-loss (\,Cross-Loss $J_3$)}}.
\end{aligned}
}
\end{equation}

\bigskip

\noindent
By replacing each side’s separate GMM with a VAE negative ELBO we obtain a fully generative row/column latent embedding.  The mutual information loss \(J_{3}\) is driven by the posterior, thereby achieving co-clustering in a deep variational mixture framework.

\section{Experiments and Results}

\subsection{Datasets and Preprocessing}

We evaluated the proposed \emph{Scalable Bayesian Co-Clustering} method on a diverse set of benchmark datasets covering both image data and web / text data. Previous results on these datasets used for benchmarking have been obtained from \cite{deepcc}. Table~\ref{tab:datasets} summarizes the key statistics for each dataset, including the number of instances (\(\#\mathrm{rows}\)), the number of features (\(\#\mathrm{columns}\)), and any known information about the label (if available). These data sets were chosen to represent different characteristics of the data. The Coil20\cite{coil20}, Yale\cite{yale}, and Fashion-MNIST\cite{xiao2017fashion} consist of image pixel intensities (relatively dense data). WebKB\cite{webkb4} and IMDb are text-based data sets that demonstrate some level of sparsity with each feature corresponding to a keyword or token presence.
As these datasets come from different domains, we apply minimal and domain-appropriate preprocessing to ensure consistent input to our co-clustering algorithm. 
For image datasets, each pixel value is scaled to be in the range \([0, 1]\). For text data, we apply the TF-IDF weighting or \(\ell_{2}\)-normalization to raw term frequency vectors.
To facilitate mini-batch training in our variational approach, we randomly shuffle the rows and columns before creating mini-batches. If necessary, we withhold a small portion of data (e.g., 10\%) as a validation set to tune hyperparameters such as learning rate or regularization coefficients.

After these steps, each data set is represented by a matrix \(n \times d\), where \(n\) is the number of instances (e.g., images, documents) and \(d\) is the number of features (e.g., pixel intensities, vocabulary terms). We feed these matrices into our co-clustering framework without further manual feature engineering.

\begin{table}[ht]
\centering
\footnotesize
\renewcommand{\arraystretch}{1.1}

\begin{tabular}{lccc}
\toprule
\textbf{Dataset} & \textbf{\#instances} & \textbf{\#features} & \textbf{\#classes} \\
\midrule
\multicolumn{4}{l}{\textbf{Image-based datasets}}\\
\midrule
Coil20               & 1440  & 1024 & 20 \\
Yale                 & 165   & 1024 & 15 \\
Fashion-MNIST-test   & 10000 & 784  & 10 \\
\midrule
\multicolumn{4}{l}{\textbf{Web-based datasets}}\\
\midrule
WebKB4               & 4199  & 1000 & 4 \\
WebKB\_cornell       & 195   & 1703 & 5 \\
WebKB\_texas         & 187   & 1703 & 5 \\
WebKB\_washington    & 230   & 1703 & 5 \\
WebKB\_wisconsin     & 265   & 1703 & 5 \\
IMDb\_movies\_keywords & 617 & 1878 & 17 \\
IMDb\_movies\_actors   & 617 & 1398 & 17 \\
\bottomrule
\end{tabular}
\caption{Summary of datasets}
\label{tab:datasets}
\end{table}
\subsection{Baseline Methods and Experimental Setup}
We compare our \emph{Scalable Robust Variational Co-Clustering} (SRVCC) method with previous co-clustering and deep clustering techniques. \textbf{SCC} \citep{dhillon2001co} is a spectral co-clustering approach that models the data matrix as a bipartite graph and partitions rows and columns by minimizing the normalized cut. \textbf{CCMod} \citep{ccmod} is a modularity-based co-clustering method that adapts Newman's modularity measure to find diagonal block structures. \textbf{DeepCC} \citep{deepcc} is a deep autoencoder-based co-clustering method that learns latent representations for rows and columns and maximizes mutual information between their cluster assignments. We also compare our results with other prior methods such as \cite{SCMK} We compare our results on the benchmark datasets with those reported in \cite{deepcc}.

\subsection{Discussion}

Tables~\ref{tab:clustering_comparison} and \ref{tab:my_table} summarize the performance of our proposed \emph{Scalable Bayesian Co-Clustering} (\textbf{SRVCC}) compared to several established baselines (e.g. SCC, SBC, CCMod, DRCC, CCInfo, SCMK, DeepCC). We report two standard metrics, clustering accuracy (\textbf{ACC}) and normalized mutual information (\textbf{NMI}), across the set of benchmarks described in the previous section.

\begin{table}[ht]
\centering
\resizebox{\textwidth}{!}{%
\begin{tabular}{ccccccccc}
\hline
Dataset & SCC & SBC & CCMod & DRCC & CCInfo & SCMK & DeepCC & \textbf{SRVCC} \\
\hline
Coil20              & $51.7 \pm 0.5$ & $66.8 \pm 1.1$ & $21.0 \pm 2.0$ & $53.2 \pm 2.4$ & $60.6 \pm 3.4$ & $65.9 \pm 0.8$ & $73.3 \pm 1.9$ & $\mathbf{72.7 \pm 2.2}$ \\
Yale                & $33.7 \pm 0.3$ & $40.0 \pm 1.3$ & $21.4 \pm 1.4$ & $13.6 \pm 0.4$ & $41.8 \pm 2.0$ & $46.6 \pm 0.5$ & $53.3 \pm 1.4$ & $\mathbf{58.1 \pm 1.7}$ \\
Fashion-MNIST-test  & $44.5 \pm 0.5$ & $45.8 \pm 0.0$ & $28.8 \pm 0.0$ & $44.1 \pm 1.8$ & $51.8 \pm 2.4$ & -             & $62.7 \pm 1.6$ & $\mathbf{68.2 \pm 1.8}$  \\
WebKB4             & $60.6 \pm 0.1$ & $47.5 \pm 0.1$ & $68.8 \pm 3.1$ & $43.6 \pm 0.4$ & $68.8 \pm 2.5$ & $52.1 \pm 0.2$ & $71.8 \pm 2.8$ & $\mathbf{83.2 \pm 1.6}$ \\
WebKB\_cornell     & $58.9 \pm 0.2$ & $54.4 \pm 0.6$ & $55.5 \pm 2.6$ & $42.6 \pm 0.0$ & $56.6 \pm 2.7$ & $49.6 \pm 0.2$ & $68.7 \pm 1.4$ & $\mathbf{74.4 \pm 2.1}$ \\
WebKB\_texas       & $59.4 \pm 0.2$ & $59.0 \pm 0.3$ & $64.5 \pm 3.0$ & $55.1 \pm 0.0$ & $64.1 \pm 3.6$ & $62.0 \pm 0.6$ & $73.8 \pm 1.2$ & $\mathbf{76.4 \pm 2.3}$ \\
WebKB\_washington  & $60.8 \pm 0.0$ & $51.7 \pm 1.0$ & $68.0 \pm 2.7$ & $46.5 \pm 0.0$ & $67.7 \pm 2.9$ & $65.4 \pm 0.4$ & $75.7 \pm 1.9$ & $\mathbf{79.3 \pm 1.2}$ \\
WebKB\_wisconsin   & $70.2 \pm 0.5$ & $72.8 \pm 1.4$ & $72.1 \pm 3.9$ & $46.1 \pm 0.0$ & $72.9 \pm 3.1$ & $73.2 \pm 0.9$ & $77.4 \pm 1.4$ & $\mathbf{81.6 \pm 2.2}$ \\
IMb\_movies\_keywords & $25.2 \pm 0.4$ & $24.0 \pm 0.2$ & $24.7 \pm 2.1$ & $12.6 \pm 1.7$ & $23.0 \pm 2.0$ & $23.3 \pm 1.1$ & $30.8 \pm 1.7$ & $\mathbf{29.3 \pm 1.1}$ \\
IMDb\_movies\_actors  & $20.5 \pm 0.4$ & $20.0 \pm 0.4$ & $20.0 \pm 1.2$ & $14.1 \pm 2.8$ & $15.6 \pm 0.7$ & $15.8 \pm 1.3$ & $23.8 \pm 0.4$ & $\mathbf{26.2 \pm 2.4}$ \\
\hline
\end{tabular}
}
\caption{Clustering accuracy comparison with SRVCC}
\label{tab:clustering_comparison}
\end{table}

\textbf{Overall Gains in Accuracy and NMI.}
SRVCC generally delivers strong co-clustering performance on most benchmark datasets. For example, on the \emph{Yale} and \emph{Fashion-MNIST} image datasets, SRVCC outperforms previous methods on both ACC and NMI, indicating that it captures more meaningful low-dimensional structures for both row (instance) and column (feature) spaces. Substantial gains are also observed for the \emph{WebKB} datasets, where SRVCC achieves significantly higher ACC compared to earlier graph-based (SCC, SBC) or modularity-driven (CCMod) techniques, and it consistently outperforms the more recent DeepCC approach on most of these web domain splits.

\textbf{Performance Variations.}
On certain benchmarks, SRVCC’s advantage is less pronounced or slightly lower than DeepCC in one of the two metrics. For example, on \emph{Coil20}, DeepCC scores slightly higher in both Accuracy (ACC) and Normalized Mutual Information (NMI). A similar trend is observed on the \emph{IMDb Keywords} split, where SRVCC’s clustering accuracy is comparable but trails DeepCC’s by a small margin, and its NMI is also slightly lower. These instances suggest that while SRVCC’s Gaussian mixture prior and scaled variational design generally improve robustness and cluster separability, domain or data-specific factors such as the intrinsic separability of the classes, the ratio of noise to signal, or the choice of hyperparameters can still pose challenges. 
\begin{figure}[ht]
    \centering
    
    \begin{subfigure}[b]{0.45\textwidth}
        \centering
        \includegraphics[width=\textwidth]{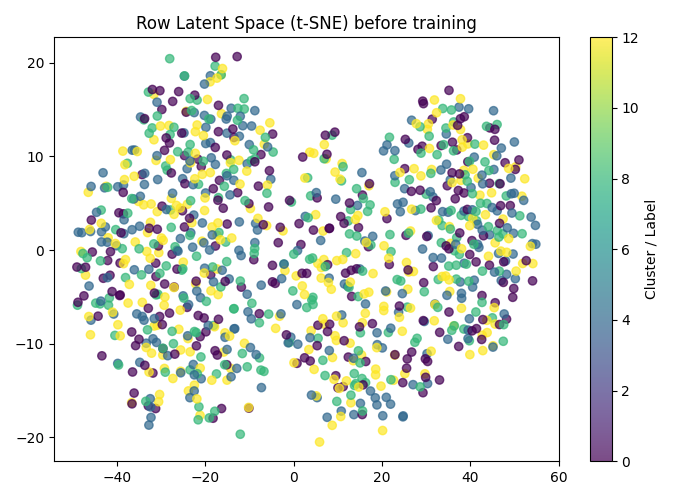}
        \caption{Initial row latent space}
        \label{fig:image1}
    \end{subfigure}
    \hfill
    \begin{subfigure}[b]{0.45\textwidth}
        \centering
        \includegraphics[width=\textwidth]{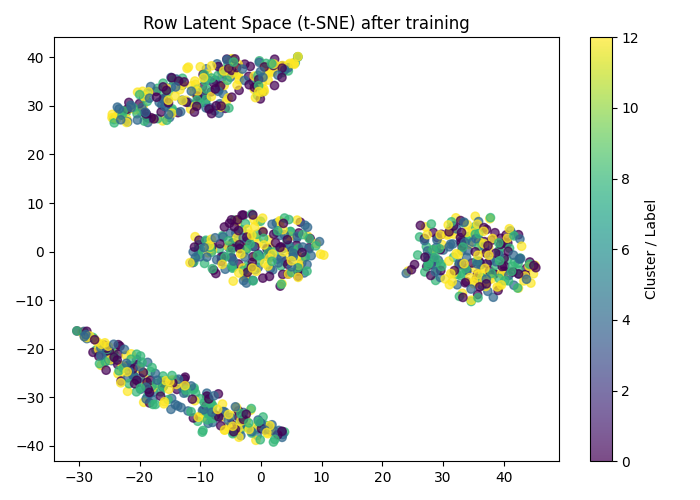}
        \caption{Final row latent space}
        \label{fig:image2}
    \end{subfigure}
    
    \vskip\baselineskip
    \begin{subfigure}[b]{0.45\textwidth}
        \centering
        \includegraphics[width=\textwidth]{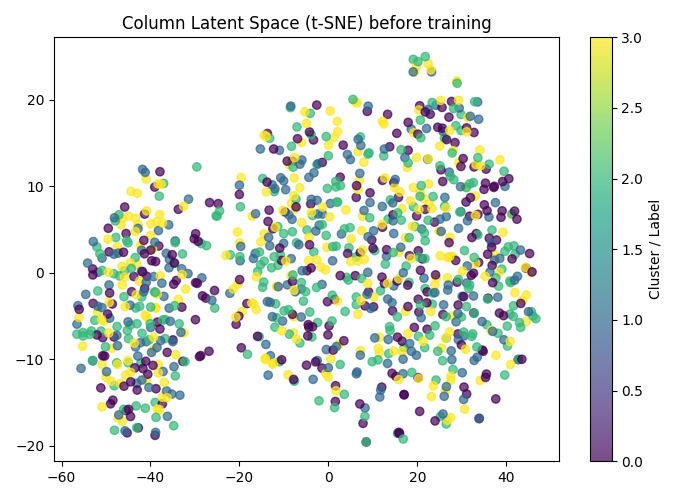}
        \caption{Initial column latent space}
        \label{fig:image3}
    \end{subfigure}
    \hfill
    \begin{subfigure}[b]{0.45\textwidth}
        \centering
        \includegraphics[width=\textwidth]{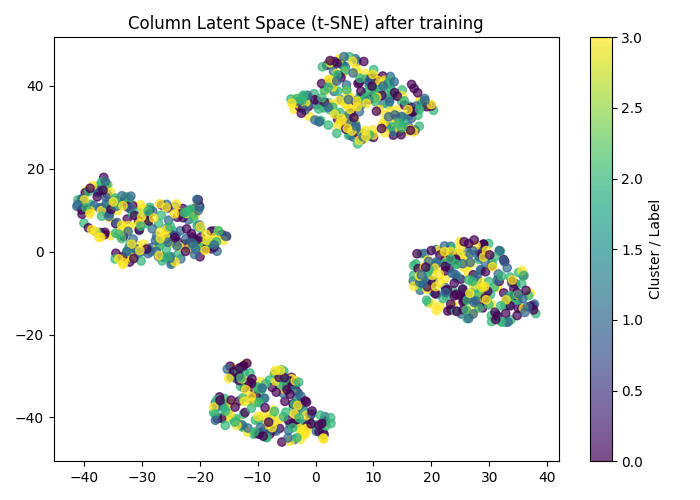}
        \caption{Final column latent space}
        \label{fig:image4}
    \end{subfigure}
    
    \caption{Visualization of latent space before and after latent space clustering on synthetic noisy data. }
    \label{fig:2x2}
\end{figure}

\begin{figure}[ht]
    \centering
    
    \begin{subfigure}[b]{0.3\textwidth}
        \centering
        \includegraphics[width=\textwidth]{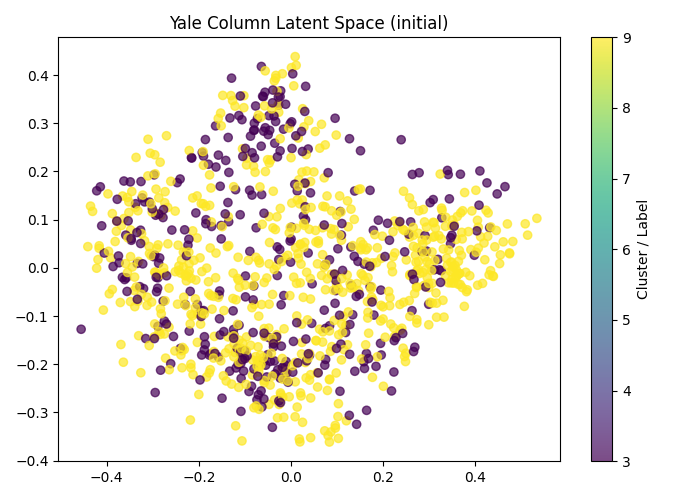}
        \caption{Intial}
        \label{fig:plot1}
    \end{subfigure}
    \hfill
    \begin{subfigure}[b]{0.3\textwidth}
        \centering
        \includegraphics[width=\textwidth]{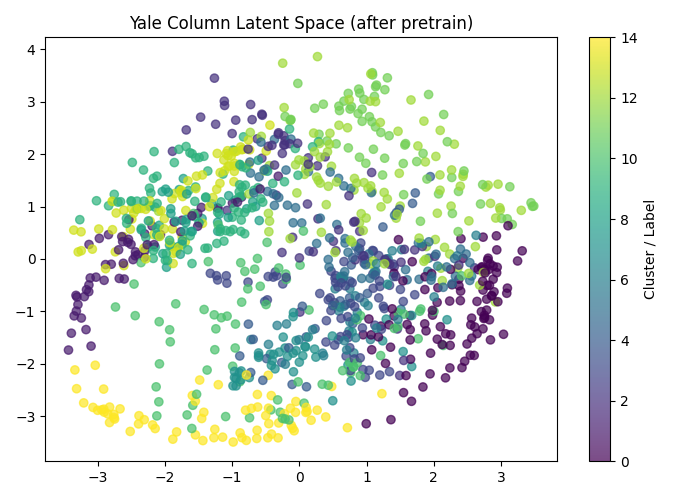}
        \caption{Post Pre-training}
        \label{fig:plot2}
    \end{subfigure}
    \hfill
    \begin{subfigure}[b]{0.3\textwidth}
        \centering
        \includegraphics[width=\textwidth]{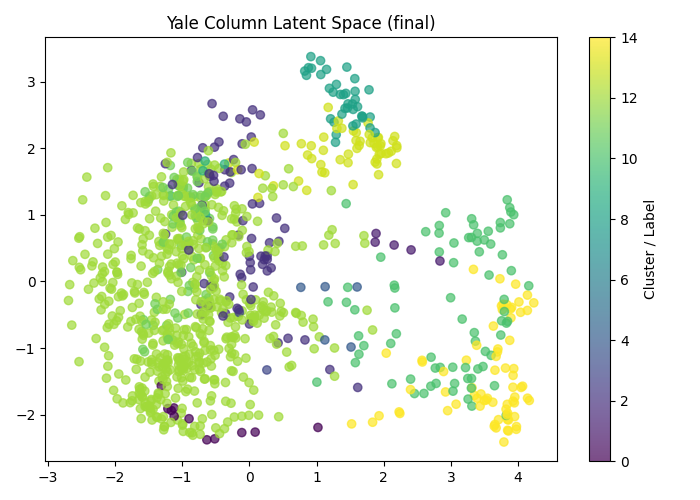}
        \caption{Final}
        \label{fig:plot3}
    \end{subfigure}

    \caption{SRVCC Feature Latent Space Visualizations on the Yale Dataset.}
    \label{fig:one-by-three}
\end{figure}

\textbf{Impact of the Scale VAE Design:}
A notable design feature of SRVCC is the use of scale parameters on the encoder’s latent means, which counteracts the tendency for posterior collapse in variational autoencoders. Empirically, this manifests in more discriminative latent embeddings. The t-SNE plots in Figures~\ref{fig:2x2} and \ref{fig:one-by-three} illustrate how the row and column embeddings become more distinctly clustered as training proceeds. Particularly on \emph{Yale} dataset, where initial latent representations are highly overlapping (Figures~\ref{fig:image1}, \ref{fig:image3}), the final learned manifolds (Figures~\ref{fig:image2}, \ref{fig:image4}) show better separation of clusters.

\begin{figure}[ht!]
    \centering
    \subcaptionbox{Co-clustering with 30\% noise level.\label{fig:img4}}{%
        \includegraphics[width=0.3\textwidth]{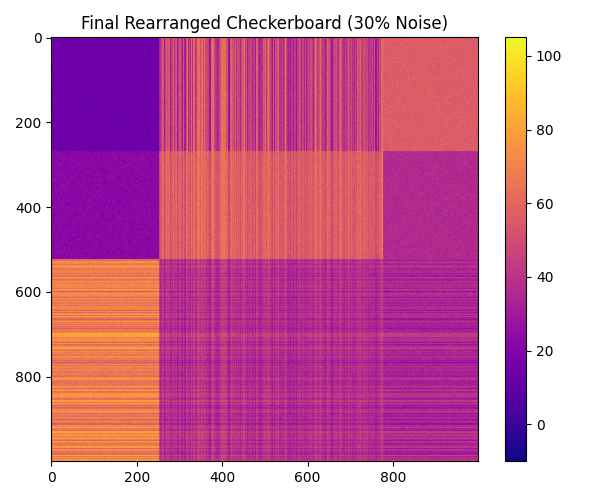}
    }\hfill
    \subcaptionbox{Co‐clustering with 50\% noise level.\label{fig:img5}}{%
        \includegraphics[width=0.3\textwidth]{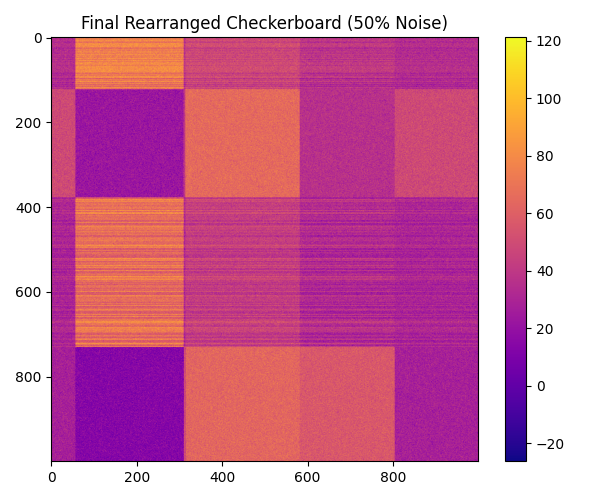}
    }\hfill
    \subcaptionbox{Co‐clustering with 70\% noise level. \label{fig:img6}}{%
        \includegraphics[width=0.3\textwidth]{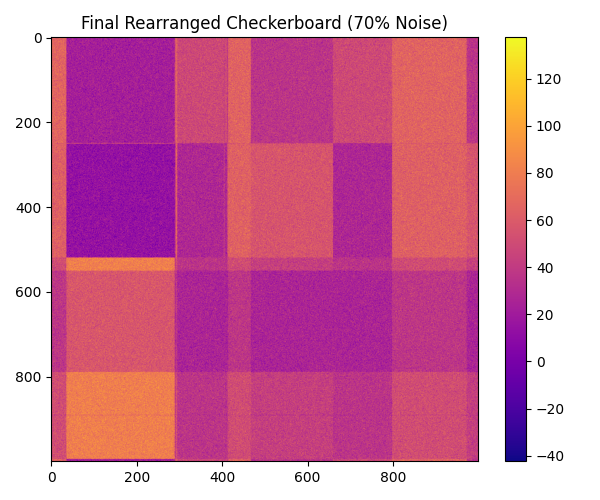}
    }

    \caption{Impact of noise levels on Co‐clustering performance: Visualization of cluster boundaries at varying noise intensities}
    \label{fig:six_images}
\end{figure}

\textbf{Robustness to Noisy Structures:}
Figures~\ref{fig:six_images} highlight that even when the original data matrix is noisy or partially shuffled, SRVCC rearranges rows and columns into coherent checkerboard co-clusters. This is especially noticeable on smaller facial-image datasets (e.g., Yale) and synthetic noisy data. The Gaussian mixture prior helps regularize the latent factors so that individual cluster components (and their variances) adapt to the underlying structure while preserving a degree of flexibility to accommodate outliers and sparse signals.

\textbf{Impact of Two stage Correcting Compositional EBLOs:} From table \ref{table2}, it is clear that the two‐stage compositional ELBO framework augmented with double reparameterized gradients (DREG) consistently yields superior accuracy and NMI across the tested datasets compared to a simple cascade of VAE’s  or feature‐only clustering scheme based Co-clustering. The dual corrections provided by two stage compositional ELBO help control noise and variance in both row/column assignments and the joint cell‐level latent space, leading to increased reliability under noisy conditions. This synergy between the Joint ELBO and the Row-Column ELBOs from the first stage ensures that each stage (Stage 1 of row‐side and  column‐side, and  stage 2 of joint) provides robust feedback to each other, thereby maintaining coherent Co‐clustering structures and enhancing overall performance.
 
\begin{table}[ht]
\centering

\resizebox{\textwidth}{!}{%
\begin{tabular}{l c c c c c c c}
\toprule
\multirow{2}{*}{\textbf{Dataset}} &
\multirow{2}{*}{\textbf{DREG}} &
\multicolumn{2}{c}{\textbf{Feature}} &
\multicolumn{2}{c}{\textbf{Simple Cascade}} &
\multicolumn{2}{c}{\textbf{Two Stage Elbo}} \\
\cmidrule(lr){3-4}\cmidrule(lr){5-6}\cmidrule(lr){7-8}
& & ACC & NMI & ACC & NMI & ACC & NMI \\
\midrule
\textbf{Fashion-MNIST-test} 
& $\times$     
  & $58.2 \pm 1.0$ & $53.5 \pm 1.2$
  & $59.0 \pm 1.4$ & $55.2 \pm 1.3$
  & $62.7 \pm 1.6$ & $60.4 \pm 0.7$ \\
\textbf{Fashion-MNIST-test} 
& $\checkmark$ 
  & $62.1 \pm 1.4$ & $58.2 \pm 1.1$
  & $64.4 \pm 1.2$ & $59.6 \pm 1.0$
  & $\mathbf{68.2 \pm 1.8}$ & $\mathbf{65.0 \pm 1.6}$ \\
\midrule
\textbf{WebKB\_washington} 
& $\times$     
  & $72.2 \pm 1.6$ & $42.8 \pm 1.3$
  & $70.6 \pm 1.5$ & $40.1 \pm 1.2$
  & $75.7 \pm 1.9$ & $45.9 \pm 1.3$ \\
\textbf{WebKB\_washington} 
& $\checkmark$ 
  & $75.4 \pm 1.3$ & $44.5 \pm 1.1$
  & $74.4 \pm 1.2$ & $43.2 \pm 1.0$
  & $\mathbf{79.3 \pm 1.2}$ & $\mathbf{48.1 \pm 1.4}$ \\
\midrule
\textbf{WebKB\_wisconsin} 
& $\times$     
  & $72.1 \pm 1.5$ & $42.6 \pm 1.2$
  & $70.6 \pm 1.3$ & $43.1 \pm 1.4$
  & $77.4 \pm 1.4$ & $46.5 \pm 1.7$ \\
\textbf{WebKB\_wisconsin} 
& $\checkmark$ 
  & $75.0 \pm 1.4$ & $46.1 \pm 1.1$
  & $73.2 \pm 1.7$ & $48.3 \pm 1.4$
  & $\mathbf{81.6 \pm 2.2}$ & $\mathbf{51.5 \pm 1.6}$ \\
\bottomrule
\end{tabular}}
\caption{Comparison among three approaches: (i) Two‐stage variational Co‐clustering  with compositional ELBOs, (ii) a simple cascade of row and column clustering encoders, and (iii) a feature‐only clustering encoder. Results using double reparameterized gradients (DREG) are also shown.}
\label{table2}
\end{table}

\begin{table}[ht]
\centering
\resizebox{\textwidth}{!}{%
\begin{tabular}{ccccccccc} 
\hline
Dataset & SCC & SBC & CCMod & DRCC & CCInfo & SCMK & DeepCC & \textbf{SRVCC} \\
\hline
Coil20              & $64.9 \pm 0.5$ & $73.9 \pm 1.1$ & $51.8 \pm 1.9$ & $65.6 \pm 2.7$ & $72.7 \pm 1.5$ & $72.5 \pm 0.9$ & $78.3 \pm 2.7$ & $\mathbf{75.0 \pm 2.1}$ \\
Yale                & $41.6 \pm 0.3$ & $49.8 \pm 1.3$ & $24.6 \pm 2.3$ & $14.2 \pm 1.2$ & $48.5 \pm 2.0$ & $49.2 \pm 1.2$ & $55.7 \pm 1.1$ & $\mathbf{61.0 \pm 1.5}$ \\
Fashion-MNIST-test  & $41.9 \pm 0.5$ & $41.3 \pm 0.0$ & $45.8 \pm 1.4$ & $42.2 \pm 1.6$ & $50.6 \pm 2.3$ & -             & $60.4 \pm 0.7$ & $\mathbf{65.0 \pm 1.6}$ \\

WebKB4              & $31.1 \pm 0.1$ & $13.0 \pm 0.1$ & $40.1 \pm 1.0$ & $31.9 \pm 1.7$ & $39.7 \pm 3.6$ & $10.0 \pm 2.3$ & $40.5 \pm 0.6$ & $\mathbf{42.3 \pm 1.2}$ \\

WebKB\_cornell      & $28.8 \pm 0.2$ & $21.0 \pm 0.6$ & $18.9 \pm 3.8$ & $11.6 \pm 0.0$ & $20.6 \pm 3.1$ & $25.7 \pm 0.5$ & $35.4 \pm 0.9$ & $\mathbf{39.3 \pm 1.8}$ \\

WebKB\_texas        & $12.6 \pm 0.2$ & $9.0 \pm 0.3$ & $16.9 \pm 2.3$ & $10.2 \pm 0.0$ & $18.2 \pm 4.4$ & $24.0 \pm 0.8$ & $42.9 \pm 1.2$ & $\mathbf{43.5 \pm 1.7}$ \\

WebKB\_washington   & $25.3 \pm 0.0$ & $9.5 \pm 1.0$ & $28.7 \pm 1.4$ & $15.7 \pm 0.0$ & $30.7 \pm 3.4$ & $30.3 \pm 0.2$ & $45.9 \pm 1.3$ & $\mathbf{48.1 \pm 1.4}$ \\

WebKB\_wisconsin    & $35.4 \pm 0.5$ & $38.2 \pm 1.4$ & $35.1 \pm 2.8$ & $20.4 \pm 0.0$ & $39.3 \pm 2.7$ & $42.9 \pm 0.4$ & $46.7 \pm 1.7$ & $\mathbf{51.5 \pm 1.6}$ \\

IMDb\_movies\_keywords  & $25.5 \pm 0.4$ & $20.6 \pm 0.2$ & $21.6 \pm 1.1$ & $6.9 \pm 0.3$ & $18.7 \pm 2.3$ & $18.4 \pm 0.8$ & $26.8 \pm 1.6$ & $\mathbf{25.3 \pm 1.2}$ \\

IMDb\_movies\_actors   & $19.3 \pm 0.4$ & $17.6 \pm 0.4$ & $14.5 \pm 0.9$ & $9.3 \pm 2.5$ & $9.7 \pm 1.0$ & $10.6 \pm 1.7$ & $20.6 \pm 2.3$ & $\mathbf{19.4 \pm 1.8}$ \\
\hline
\end{tabular}
} 
\caption{NMI Clustering results across various datasets.}
\label{tab:my_table}
\end{table}



\begin{figure}[ht]
    \centering
    \begin{subfigure}[b]{0.45\textwidth}
        \includegraphics[width=\textwidth]{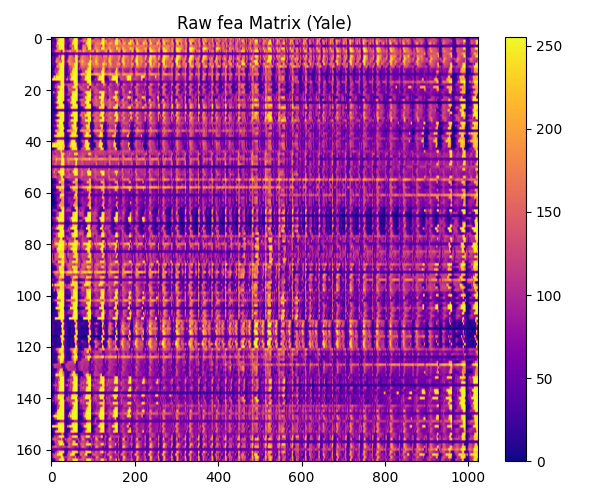} 
        \caption{Input Data Matrix from Yale Dataset}
        \label{fig:sub1_}
    \end{subfigure}
    \hfill
    \begin{subfigure}[b]{0.45\textwidth}
        \includegraphics[width=\textwidth]{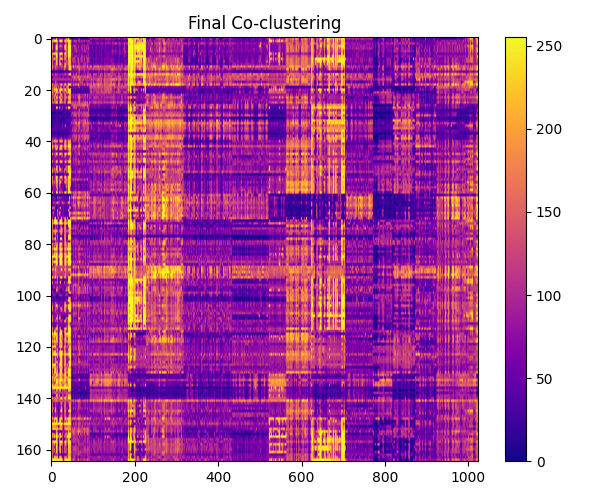} 
        \caption{Final Co-clusters (15 Clusters Detected)}
        \label{fig:sub2_}
    \end{subfigure}

    \caption{Plots showing the Inital input data matrix and final rearranged Coclsuter Checkerboard on the Yale Dataset.}
    \label{fig:two_side_by_side_1}
\end{figure}

\section{Conclusion}
In this paper, we present a fully variational Co‐clustering framework that directly learns row and column clusters within the latent space, thereby eliminating the need for a separate clustering step. Using a doubly reparameterized ELBO, our approach improves gradient signal‐to‐noise separation and naturally aligns latent modes with row and column clusters via a Gaussian Mixture Model (GMM) prior for both instances and features. Furthermore, the regularized end-to-end noise learning architecture jointly reconstructs data and handles corrupted or missing inputs through the KL divergence, promoting robustness under high‐dimensional or noisy conditions. To counteract posterior collapse, we used a scale modification that strictly increases the encoder’s latent means in the reconstruction pathway, preserving richer latent representations without inflating the KL term. Finally, we introduced a mutual information-based cross-loss to ensure coherent clustering of rows and columns. Empirical studies on diverse real-world datasets, spanning numerical, textual, and image-based domains, demonstrated that our method consistently outperforms state‐of‐the‐art alternatives, achieving greater accuracy and resilience to noise while maintaining the advantages of prior Co‐clustering approaches.

Our future research direction focuses on extending our co-clustering approach to biomedical domains, particularly for early biomarker discovery and longitudinal analyses of Parkinson's disease (PD) using Multimodal data sources. Preliminary investigations into integrating imaging and clinical biomarkers for early PD biomarker discovery are provided in the appendix.
\section*{Acknowledgement}
This research was supported in part by a grant from NIH DK129979, in part by the Peter O’Donnell Foundation, the Michael J. Fox Foundation, and the Jim Holland–Backcountry Foundation.

\appendix
\section{Appendix}
 \subsection{PPMI dataset}

 We apply our co‐clustering approach to a clinical/imaging dataset from the Parkinson’s Progression Markers Initiative (PPMI). Each patient is represented by standard PD measures (e.g., UPDRS Part III for motor symptoms, MoCA for cognitive screening, QUIP for impulse control disorder risk) and imaging biomarkers (T1‐MRI volumetrics of key subcortical regions, DTI metrics indicating white‐matter integrity, and DaTSCAN for dopamine transporter binding). Co‐clustering simultaneously partitions the patient dimension by symptom severity or disease manifestation and the biomarker dimension by highly covarying neural features. This yields interpretable blocks of (\emph{patient subgroups} \(\times\) \emph{brain‐imaging features}) that reveal distinct PD subtypes. For instance, striatal DaTSCAN and volumetric measures correlate strongly with higher UPDRS‐III motor scores, while limbic/cortical biomarkers co‐cluster with more severe cognitive decline. Overall, these findings demonstrate an integrated view of which neural and clinical factors differentiate PD progression pathways. Our PPMI data input and induced co-clusters are shown in Fig.\ref{p2}.

\subsection{Visualizations}

\begin{figure}[ht]
    \centering
    \captionsetup{justification=centering} 
    \begin{minipage}{0.32\textwidth}
        \centering
        \includegraphics[width=\textwidth]{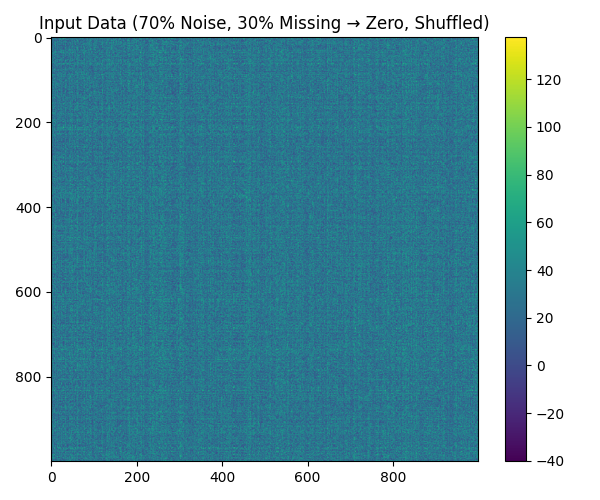}
        \caption{Input data with 70\% noise and 30\% missing values (shuffled).}
    \end{minipage}
    \begin{minipage}{0.32\textwidth}
        \centering
        \includegraphics[width=\textwidth]{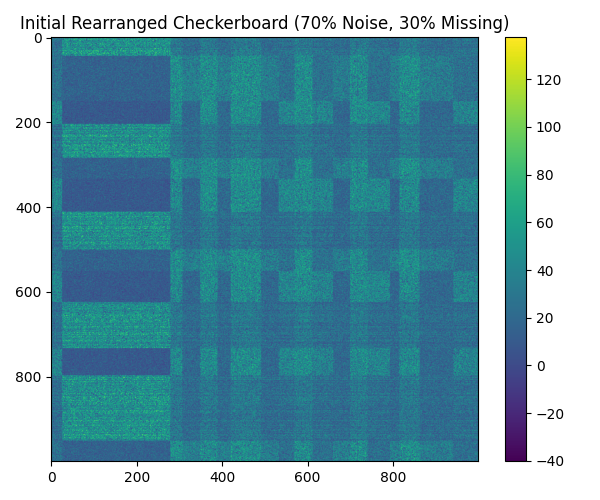}
        \caption{Initial co-clustering assignments before model training.}
    \end{minipage}
    \begin{minipage}{0.32\textwidth}
        \centering
        \includegraphics[width=\textwidth]{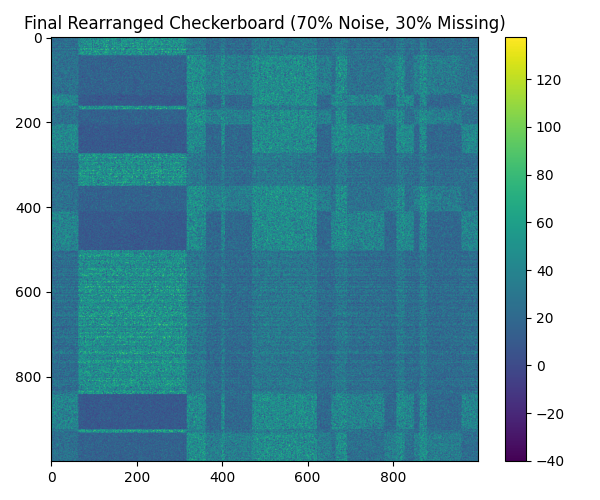}
        \caption{Final co-clustering assignments after model training.}
    \end{minipage}
    \caption{Visualization of SRVCC Co-clustering process on noisy and incomplete data.}
    \label{p1}
\end{figure}

\begin{figure}[ht]
    \centering
    \begin{subfigure}[b]{0.45\textwidth}
        \includegraphics[width=\textwidth]{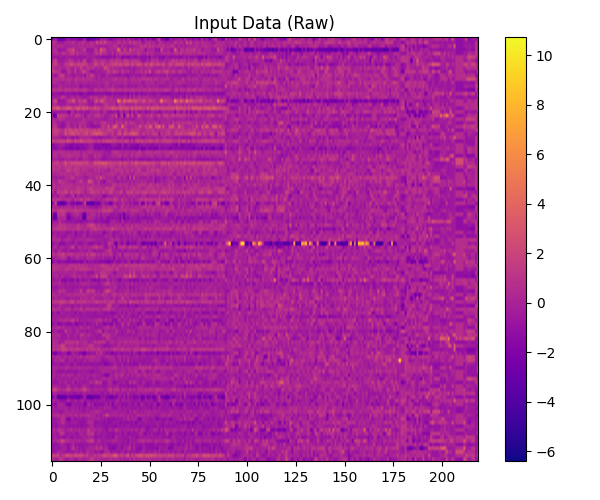} 
        \caption{Input Data from the PPMI Dataset}
        \label{fig:sub1a}
    \end{subfigure}
    \hfill
    \begin{subfigure}[b]{0.45\textwidth}
        \includegraphics[width=\textwidth]{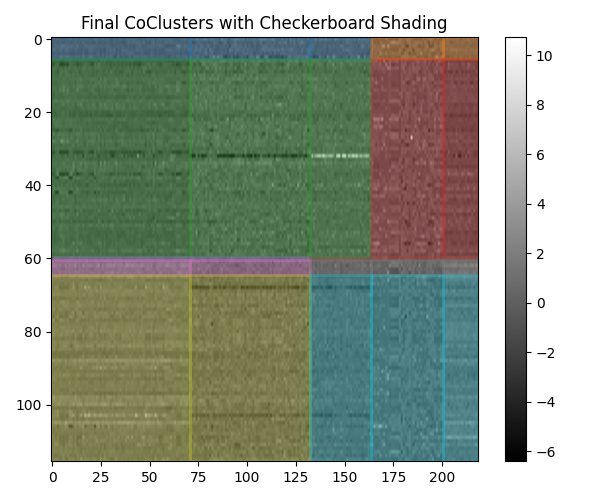} 
        \caption{Final Co-clusters detected}
        \label{fig:sub2a}
    \end{subfigure}

    \caption{Plots showing the Inital normalized input data matrix and final rearranged Coclsuter Checkerboard on the PPMI Dataset.}
    \label{p2}
\end{figure}

We present additional visualizations of the co-clustering process across different datasets. Figure~\ref{p1} illustrates the co-clustering pipeline on a dataset with 70\% noise and 30\% missing values, showing the input data, initial co-clustering assignments, and final learned co-clusters. Figure~\ref{p2} demonstrates the process on the PPMI dataset, comparing the initial shuffled input with the final rearranged co-cluster structure. Similarly, Figure~\ref{p3} provides an analogous visualization for the WebKB4 dataset, highlighting the transformation from raw input to structured co-clusters. These results further support the robustness and adaptability of our scalable Bayesian co-clustering approach in handling diverse data modalities and noise levels.

\begin{figure}[ht]
    \centering
    \begin{subfigure}[b]{0.45\textwidth}
        \includegraphics[width=\textwidth]{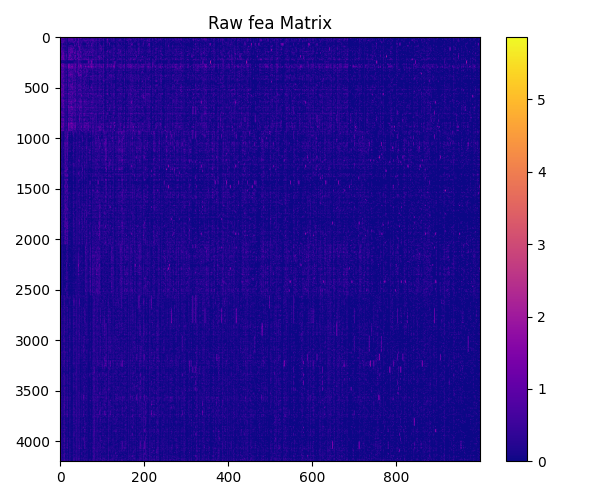} 
        \caption{Input from the WEbKB4 Dataset}
        \label{fig:sub1b}
    \end{subfigure}
    \hfill
    \begin{subfigure}[b]{0.45\textwidth}
        \includegraphics[width=\textwidth]{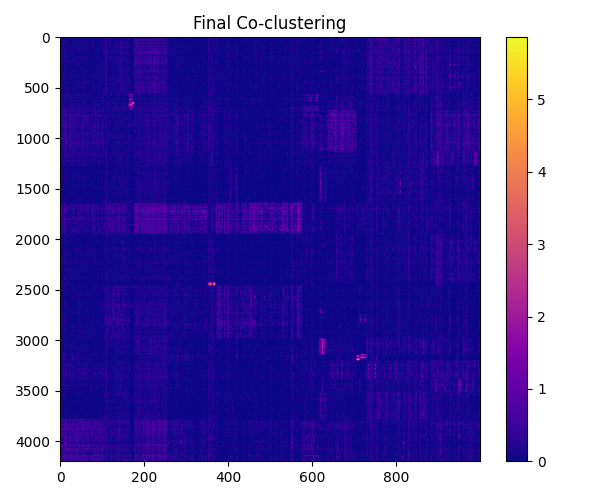} 
        \caption{Final Co-clusters detected}
        \label{fig:sub2b}
    \end{subfigure}

    \caption{Plots showing the Inital input data matrix and final rearranged Coclsuter Checkerboard on the WEbKB4 Dataset.}
    \label{fig:two_side_by_sidec}
\end{figure}
    \label{p3}

\section{Discussion}

In this section we discuss our findings that emerged from the co-clustering of the PPMI dataset, which includes imaging features (DaTSCAN and volumetric measures) and clinical scale scores (e.g., MOCA, UPDRS, SCOPA). We obtained 10 co-cluster groups of which we have 4 Patient groups labeled \textbf{0}--\textbf{4}.

\subsection{Cluster 2: A Singleton (or Near-Singleton) with Mild Disease}

One of the most striking findings is that \texttt{Cluster 2} frequently has the \emph{same exact min and max} for many variables. For example \texttt{DaTSCAN} measures (e.g.,  \texttt{DaTSCAN-Left-Lateral-Ventricle} has min = 0.243315969 and max = 0.243315969).
\texttt{Volume} measures (e.g., ``Volume-Left-Lateral-Ventricle'' min = max = 24187).
 \texttt{Clinical scores} (MOCA = 29, UPDRS-3 = 6, UPDRS total = 14).
This indicates that \texttt{Cluster 2} is essentially a single subject (or a very small number of nearly identical patient sets). Where on average, patient groups have \texttt{UPDRS-3} = 6 (mild motor severity), \texttt{MOCA} = 29 (high cognitive functioning), \texttt{HY stage} = 1, \texttt{BMI} $\approx$ 25.4, \texttt{Age at Diagnosis} $\approx$ 60.
All these values suggest a relatively mild, early-stage PD profile with preserved cognition and low symptom burden. Cluster 2 can be interpreted as an ``outlier'' or a very distinct mild case of PD \cite{Batla2014}\cite{Hoops2009}.

\subsection{Cluster 3: Generally Older, With Relatively High DaTSCAN Ratios, But Variable Motor Severity}
\noindent
\\
\textbf{Age at Diagnosis and Disease Onset: }
Cluster 3’s \texttt{agediag} median is around 66.4, which stands out compared to the 60--64 range in other clusters. Even the minimum and maximum ages at diagnosis extend over a wide range (e.g., from early 50s into the 70s), but the older median underscores that most individuals in this group are indeed on the higher end of the age spectrum \cite{Johansson2024}\cite{Kaasinen2017}. 

\noindent
\\ 
\textbf{Dopaminergic Uptake Across Multiple Regions:}
While \texttt{DaTSCAN Left Lateral Ventricle} is sometimes seen as a reference or background measure, other ratios such as \emph{Left/Right Putamen} or \emph{Left/Right Caudate} also appear higher in Cluster 3 than in most other clusters. Older PD patients tend to have lower dopaminergic binding, so the relatively preserved DaTSCAN signals in Cluster 3 might suggest a subgroup whose striatal dopaminergic pathways are less affected, or who exhibit individual biological variation in preserving dopaminergic function longer \cite{Matesan2018}\cite{Stoessl2017}.
We also see cluster‐specific medians for putamen/caudate that exceed 1.2 or 1.3 (whereas other clusters have medians around 0.8--1.0).

\noindent
\\ 
\textbf{Motor Severity (UPDRS-3) Range: }
UPDRS‐3 \cite{Goetz2008} has a large range in Cluster 3: roughly from 4 to 36. The median of 16.5--18 places it in a moderate severity tier, typical of HY stage 2--2.5 PD. A closer split reveals some individuals scoring under 10 (mild) and others in the 30s (more advanced). Even those with higher motor scores often show DaTSCAN values that remain comparatively higher than expected for that symptom level.

\noindent
\\ 
\textbf{Cognitive Status (MOCA): }
The median MOCA \cite{Hoops2009_MoCA} of about 26.5 is slightly below the 27--28 seen in some clusters, indicating a modest decline but not severe impairment.
The data still show a broad spread (low 20s up to near 30), reflecting a mix of mild cognitive changes in these older participants.

\noindent
\\ 
\textbf{Possible Sub‐Phenotypes: }
The combination of relatively preserved dopaminergic imaging, moderate motor range, and older age could indicate a slower‐progression subtype of PD or a phenotype where non‐dopaminergic pathways are more influential for motor symptoms.
 Another possibility is a small cohort of older individuals whose DaTSCAN values remain at the higher end compared to age‐matched peers, pointing to biological resilience in dopaminergic terminals despite advanced age.

\noindent

Hence we conclude that \emph{Cluster 3} emerges as an older group displaying a balance of \emph{maintained} striatal DaTSCAN signals yet moderate and wide-ranging UPDRS-3 scores. This nuanced relationship underscores how age, dopaminergic imaging, and clinical severity in PD do \emph{not} always follow a simple, linear pattern \cite{lorio2019combination}. It highlights a more complex or heterogeneous disease process, suggesting further division into subgroups might be warranted for deeper phenotypic classification.

\subsection{Similarities and Differences in Cluster 0 and Cluster 1}
\textbf{MRI Volumes and Spread: }
 Volumetric measures (e.g., \texttt{Volume-Left- Lateral- Ventricle} mean $\approx 23{,}197$, std $\approx 13{,}776$) indicate a moderate degree of atrophy but not extreme.
 Many other structures show midrange volumes with lower standard deviations, implying a \emph{more homogeneous} subset in MRI terms. In Cluster 1 Overall, volumes have larger standard deviations. For instance, \texttt{Volume-Left-Lateral-Ventricle} might vary from $\sim 5{,}438$ to nearly $48{,}379$. This wide span mixes some very mild (small ventricle) and some advanced (high ventricle) atrophy profiles.
    \noindent
Hence, \textbf{Cluster 1} lumps together extremes, while Cluster 0 is narrower in range yet still moderately affected.

\noindent
\\
\textbf{ Cognitive Scores (MOCA): }
In Cluster 0, The Median \textbf{MOCA} $\approx 28.5$, among the highest for multi-subject clusters, implying mostly preserved cognition.
Whereas in Cluster 1: The Median $\approx 27$ and mean $\approx 26.4$, still borderline normal but slightly lower on average than Cluster 0. 
 Some individuals drop to mid-low 20s while others remain near 29--30.

\noindent
Hence, Cluster 0 seems more uniformly high in MOCA, whereas Cluster 1 has both unimpaired and mildly impaired cognition.

\noindent
\\
\textbf{ Motor Severity (UPDRS-3): }
For Cluster 0, the median UPDRS-3 $\sim 16.5$--20, firmly ``moderate'' but not extremely broad.
And Cluster 1, the median $\approx 17.5$ but a large range (7 to 38), indicating both near-mild and more advanced individuals.
\noindent
Thus, both clusters indeed represent ``moderate PD,'' but Cluster 1 includes a wider motor-severity spectrum.
\textbf{Age,  BMI and Outliers: }
For Cluster 0 median AgeDiag (age) is  $\approx 60.5$, while Cluster 1's median $\approx 64.1$, making 1 older on average.
The BMI of  Cluster 1 features a standard deviation which is on the higher end, and so is the median relative to Cluster 0.

\noindent
Hence Cluster 0 may represent a \emph{homogeneous, midlife} PD cohort with decent cognition (high MOCA) and moderate, stable motor severity. MRI and DaTSCAN also exhibit moderate, less variable values.
Whereas, Cluster 1 comprises an \emph{older} and more varied PD population. Many appear moderate, yet the broader standard deviations and outliers suggest some are quite mild or quite advanced. This group likely contains multiple sub-profiles in a single cluster.

Overall, both are moderate PD, but Cluster 1 is older on average and more diverse in its imaging and clinical expression, whereas Cluster 0 is younger, more cognitively intact, and more uniform across different measures.

\subsection{Cluster 4:  Broad Group with Intermediate Severity}

Many volumetric and DaTSCAN metrics in Cluster 4 are not extreme but show broad ranges.For Example, \texttt{Volume-Left-Lateral-Ventricle} median is 13975, mean 16064, with a wide range. The DaTSCAN LeftCaudate median $\approx 1.28$, an intermediate value relative to 1 or 3.
The clinical feature values include \texttt{MOCA} with a median of $\approx 27$,
\texttt{UPDRS-3} median of $\approx 19$ and 
\texttt{Age at diagnosis} median $\approx 61.4$.
 
Hence, Cluster 4 also has moderate severity, similar to Cluster 1 but with slightly different imaging and younger median age.

\subsection{Discussion}
Cluster 2 contains a single individual whose mild PD symptoms (low UPDRS), high MOCA, and normal imaging features set them apart as distinct. Clusters~0, 1, and~4 share a “moderate PD” profile with variations in average ages and cognitive scores (with Cluster~0 exhibiting the best cognition), while Cluster~3 comprises an older group that spans a wide motor range and unexpectedly demonstrates higher DaTSCAN values. Imaging observations reveal that left/right caudate and putamen ratios are highest in Cluster~3, with Clusters~1 and~4 displaying intermediate values; the individual in Cluster~2 also appears mild or near-normal in imaging.

feature dimensions.

\begin{figure*}[htbp]
  \centering
  \includegraphics[width=\textwidth]{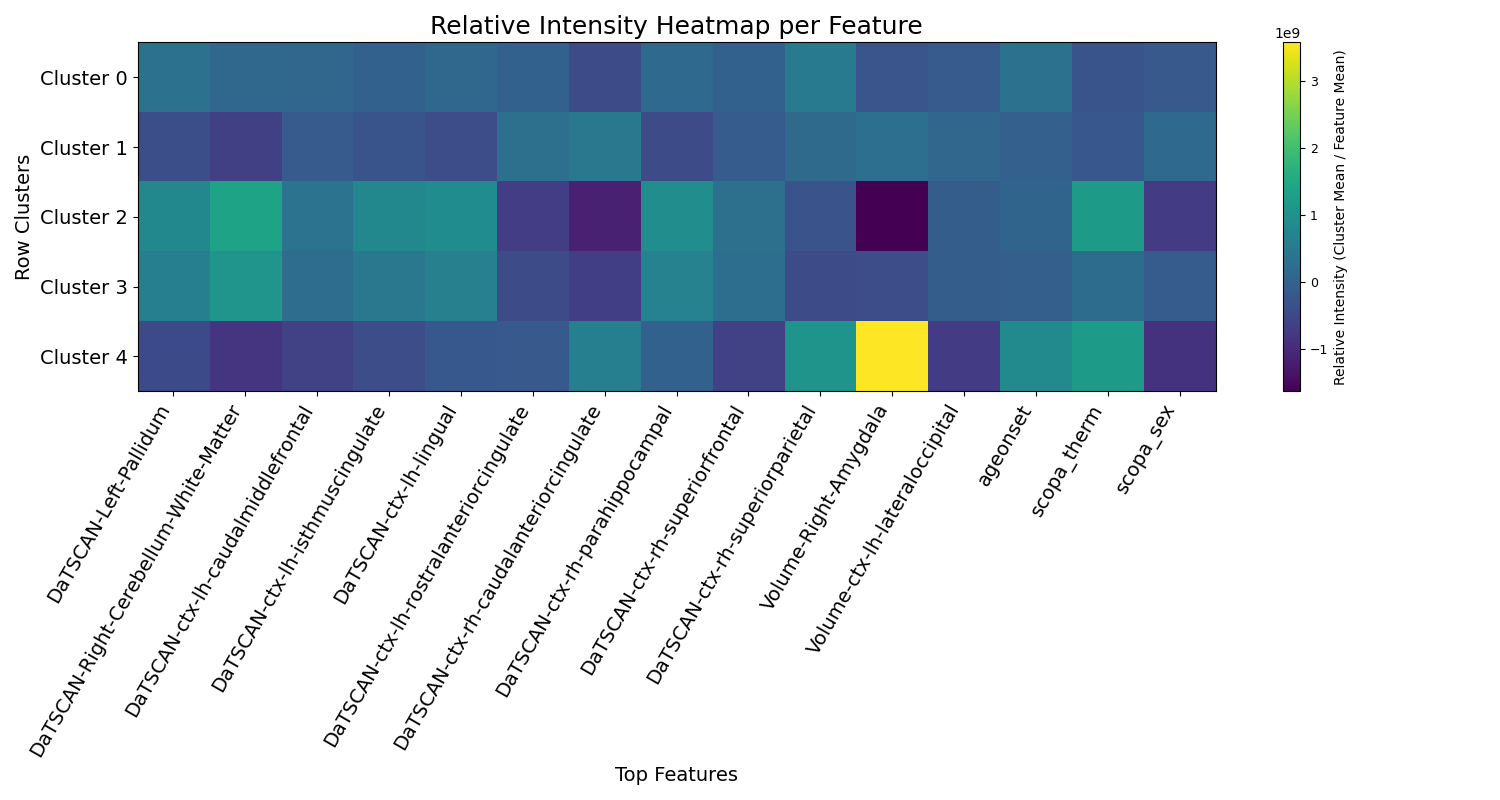}
  \caption{Feature relative heatmap of the top 15 features with highest inter group difference.}
  \label{fig:feat}
\end{figure*}

A partial correlation between volume and cognition is observed, as Cluster~0 shows robust volumes alongside high MOCA scores. However, the relationship is not strictly linear—Cluster~3, for example, is older yet does not consistently exhibit small striatal volumes. This underscores the multifactorial nature of these profiles.

Overall, our co-clustering yields a single “mild” outlier (Cluster~2), several broad moderate clusters (0, 1, 4) that differ in cognition, age, or data spread, and an older group (Cluster~3) with a wide motor range and higher dopaminergic imaging signals. This suggests that imaging, age, and clinical severity do not always align in a purely linear fashion, highlighting distinctive phenotypes within the PPMI population.

\section{Potential biomarkers}

From the cluster-level patterns, we see that a set of imaging and clinical variables consistently stand out as distinguishing features across clusters. These variables appear to capture meaningful differences in disease stage (mild vs.\ moderate), cognitive performance, or structural/neurochemical changes. Thus, they show promise for being potential biomarkers in Parkinson’s disease (PD).

\noindent
\\
\textbf{Dopaminergic Imaging (DaTSCAN) Metrics: }
\noindent In dopaminergic imaging (\texttt{DaTSCAN}) metrics, measures of striatal uptake particularly \texttt{DaTSCAN-Left-Putamen}, \texttt{DaTSCAN-Right-Putamen}, \texttt{DaTSCAN-Left-Caudate}, and \texttt{DaTSCAN-Right-Caudate} — are known from Parkinson’s disease (PD) literature to be highly relevant, as dopaminergic loss typically begins in the dorsal-lateral putamen. Differences in these DaTSCAN ratios across clusters (including the presence of a “super mild” outlier with near-normal striatal uptake) suggest that these measures can robustly differentiate milder from more moderate PD phenotypes. At the same time, other subcortical regions, such as \texttt{DaTSCAN-Left/Right-Pallidum} and \texttt{DaTSCAN-Left/Right-Thalamus}, may also be altered; indeed, certain clusters (e.g., an older cluster labeled “Cluster 3”) displayed relatively higher or less reduced uptake in these regions than expected, indicating that these additional metrics may also help characterize PD subtypes. In short, \texttt{relative striatal DaTSCAN} signal (particularly in the putamen) remains a leading candidate biomarker for PD severity or subtype.

\noindent
\\
\textbf{Structural (Volumetric) MRI Measures: } In MRI imaging, \texttt{ventricular volumes} \cite{Dalaker2011}\cite{Camicioli2011} often serve as markers of global atrophy or disease burden, with the cluster analysis revealing substantial between-cluster differences (ranging from approximately 10,000,mm\textsuperscript{3} to 40,000–50,000,mm\textsuperscript{3}). These larger ventricular measures typically correlate with older age or more advanced neurodegeneration. Meanwhile, \texttt{striatal volumes} though secondary to dopaminergic depletion in PD—can show structural changes over time that distinguish different patient clusters, especially when combined with \texttt{DaTSCAN} signals. Various cortical volumes (e.g., frontal or temporal regions) also vary across clusters, though in typical PD (as opposed to some atypical parkinsonian syndromes), cortical atrophy is usually less pronounced in earlier stages, whereas subcortical and ventricular alterations are more consistent. Overall, \texttt{enlarged ventricles} and \texttt{reduced subcortical volumes} (putamen, caudate, etc.) emerge as strong imaging features that help differentiate PD subgroups.

\noindent
\\
\textbf{Clinical \& Neuropsychological Measures: } Standard clinical measures of motor involvement in PD, notably \texttt{UPDRS3\_score}, \texttt{UPDRS\_totscore}, and \texttt{HY stage}, provide reliable distinctions between subgroups \cite{Lewis2005}\cite{Marras2020}, as evidenced by the cluster analysis showing that patients with lower UPDRS3 scores and HY stages (e.g., one mild outlier at UPDRS3=6 and HY=1) clustered separately from those with moderate scores (UPDRS3 medians of 16--20 and HY around 2--3). Cognitive assessments, such as the \texttt{MOCA} test, are similarly discriminative, with higher values (28--29) distinguishing "mild/well‐preserved" individuals from those at moderate (26--27) or lower levels of cognitive function. Meanwhile, other nonmotor scales, including \texttt{SCOPA} (sleep and autonomic domains), \texttt{GDS} (depression), \texttt{STAI} (anxiety), and \texttt{HVLT} (memory), add important nuance to the overall clinical picture, though they can be more variably affected in PD. Certain \texttt{SCOPA} subdomains (such as SCOPA-AUT) or the \texttt{Epworth Sleepiness Scale} (\texttt{ESS}) may track distinct PD phenotypes but generally show lower predictive consistency for disease progression compared to motor severity or dopaminergic imaging markers.

\bibliographystyle{acm}
\bibliography{refs}
\end{document}